\documentclass[letterpaper,10pt,conference]{ieeeconf}
\IEEEoverridecommandlockouts

\usepackage{mathtools}
\usepackage{amsmath,mleftright}
\usepackage{xparse}
\usepackage{comment}
\usepackage{amssymb}

\usepackage{graphicx}
\let\labelindent\relax
\usepackage{enumitem}
\usepackage{tabularx}
\usepackage[table]{xcolor}
\usepackage[colorlinks=true,urlcolor=blue]{hyperref}  
\usepackage[noadjust]{cite}
\usepackage{subcaption}
\usepackage[scr=dutchcal]{mathalpha}
\usepackage{array}
\usepackage{tabularx}

\newtheorem{lemma}{Lemma}%

\newtheorem{proposition}{Proposition}

\newtheorem{definition}{Definition}%

\hypersetup{
    colorlinks=true,
    linkcolor=darkgray,
    filecolor=magenta,      
    urlcolor=cyan,
    pdftitle={Overleaf Example},
    pdfpagemode=FullScreen,
     citecolor    = darkgray
    }

 	\definecolor{newgreen}{RGB}{0,127,0}
  \definecolor{newblue}{RGB}{0,0,127}

\newcolumntype{Y}[1]{%
    >{\hsize=#1\hsize
      \linewidth=\hsize
      \centering\arraybackslash}X%
}

\usepackage{titlesec}

\usepackage{xcolor} 

\begin{document}
\newcommand{\extt}[1]{\textcolor{magenta}{#1}}

\title{$h$-NESDF: An Infinitely Differentiable Signed Distance
Between Two Sets with Robotic Applications}

\author{
    Vinicius Mariano Gonçalves,
    Pedro Antonio de Souza Campos Rodrigues%
    \thanks{V. M. Gonçalves is with the Department of Electrical
    Engineering, Federal University of Minas Gerais (UFMG),
    Belo Horizonte, Brazil.
    Email: \texttt{marianoauto@ufmg.br}. P. A. S. C. Rodrigues with the Control and Automation Engineering Undergraduate Program at Federal University of Minas Gerais (UFMG),
    Belo Horizonte, Brazil. E-mail: \texttt{pedro.ufmg2@gmail.com}}
}

\maketitle

\begin{abstract}
Computing distances between sets is important in many robotic applications, such as motion planning and control. In many applications, derivatives of the distance are needed as well. It is well known that the Euclidean distance is not always differentiable, which create issues to algorithms that need a derivative. In this paper, we introduce NESDF, the Normal Euclidean Signed  Distance Function between two general (with mild regularity assumptions but possibly non-convex) sets. This metric is still not differentiable, but it allows us to introduce a generalization, $h$-NESDF, which has a smoothing parameter $h$ such that the metric is infinitely differentiable for all $h>0$ and that reduces to the NESDF when $h$ goes to $0$. The metric is amenable to GPU parallelization, and thus can be computed efficiently. We show extensive simulation results to evaluate its viability in motion planning.
\end{abstract}


\section{Introduction}
Distance computations are central to robot motion planning, collision detection, and control. Their integration into gradient-based planners and feedback controllers often requires more than accurate distance values: the underlying functions must also provide sufficiently regular derivatives. Control Barrier Functions (CBFs), for example, rely on gradients, and some formulations require higher-order derivatives of distance-like functions to enforce safety constraints \cite{Singletary2021,Singletary2022,Thirugnanam2022cbf,Thirugnanam2022,Thirugnanam2023,11060906,Ferreira2026}. Insufficient regularity can lead to excessively large control inputs \cite{Goncalves2024} or noisy control signals \cite{Goncalves2026}. Motion planners such as CHOMP \cite{Ratliff2009} similarly use distance gradients to update trajectories and resolve collisions during optimization. In this setting, signed distance-like functions are particularly useful because their negative values encode overlap and provide information about infeasible configurations.

However, Euclidean distance functions are generally not differentiable at every configuration \cite{Escande2014}. A common source of nonsmoothness is a change in the geometric features that determine the closest points. Such transitions often involve nonunique \emph{witness points}, namely, pairs of points, one on each object, that attain the minimum distance. This can produce abrupt changes in the distance gradient as the relative pose of the objects varies.

Several works have investigated differentiable alternatives to Euclidean distance functions for robotic applications. The approach in \cite{Goncalves2024} introduced an infinitely differentiable, unsigned distance-like function between convex sets and applied it to second-order kinematic control. The method generalized von Neumann's alternating projection algorithm into an iterative distance computation, but remained cumbersome to evaluate and did not guarantee a zero value when the sets overlapped. Subsequently, \cite{Goncalves2026} introduced a substantially simpler formulation that correctly vanishes upon overlap, while retaining the convexity assumption and unsigned nature of the function. The work in \cite{Nunes2026} developed a differentiable distance-like function between a point and an object represented as a set of discrete points, and used it to construct safe and smooth guiding vector fields that avoid collision with obstacles represented as point clouds. Signed differentiable distance metric between convex sets were also developed in \cite{11060906,bartelt2026h}, using different smoothing techniques and demonstrating applications to CBF-based control. 
Related constructions also arise outside robotics, particularly in surface reconstruction. For example, \cite{9578427} reconstructs a smooth approximation of a point-to-set signed distance field from sampled surface data. Similarly, \cite{hicken2022explicit} proposes an explicit differentiable level-set formulation that approximates the signed distance to a geometry. 
The methods discussed above either assume convexity for both extended objects or address point-to-set distances. More recently, \cite{beker2025smooth} introduced a smooth signed distance metric for nonconvex objects aiming to use it within a contact simulation framework. Table \ref{tab:comparison} shows comparisons between different differentiable distance approaches.

\begin{figure}[!t]
    \centering
    \includegraphics[width=0.5\textwidth]{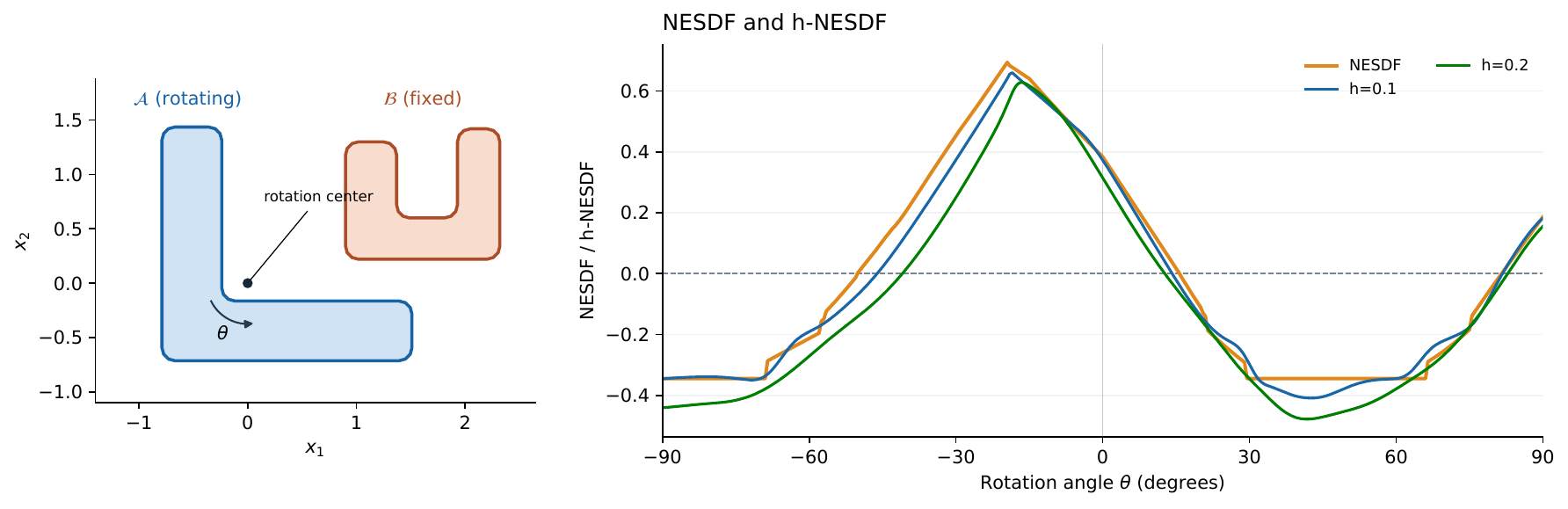}
    \caption{The proposed NESDF and $h$-NESDF distance metrics between two objects $\mathcal{A}$ and $\mathcal{B}$ when one of them is rotating. Note that the NESDF has some sharp corners (non-differentiabilities), whereas the $h$-NESDF is infinitely differentiable.}
    \label{fig:nesdf_comparison}
\end{figure}


\begin{table*}[t]
    \centering
    \scriptsize
    \renewcommand{\arraystretch}{1.2}
    \renewcommand{\tabularxcolumn}[1]{m{#1}}

    \caption{Comparison of differentiable distances proposed
    in the literature. Here, ``Detect Collision'' means that
    the differentiable distance is positive if and only if
    there is no overlap between the objects, and $h$ is a smoothing parameter.}
    \label{tab:comparison}

    \begin{tabularx}{\textwidth}{|
        Y{0.8}|  
        Y{1.1}|  
        Y{0.4}|  
        Y{1.2}|  
        Y{1.5}|} 

        \hline
        \textbf{Work}
        & \textbf{Type of Set}
        & \textbf{Signed}
        & \textbf{Detect Collision}
        & \textbf{Time Complexity} \\
        \hline

        Liu et al. (2021) \cite{9578427}
        & A point and a possibly nonconvex set
          in $\mathbb{R}^3$, represented as a point cloud
        & Yes
        & No
        & $\mathcal{O}(K)$ per query for the full formula, in which $K$ is the number of points.
          Local evaluation uses $k=10$ nearby points,
          plus neighbor-search cost \\
        \hline

        Hicken et al. (2022) \cite{hicken2022explicit}
        & A point and a possibly nonconvex set
          in $\mathbb{R}^2$ or $\mathbb{R}^3$, represented as a point cloud
        & Yes
        & No
        & $\mathcal{O}(K)$ per query for the full formula, in which $K$ is the number of points.
          The implementation uses $k=10$ nearest points,
          plus neighbor-search cost \\
        \hline

        Gonçalves et al. (2024) \cite{Goncalves2024}
        & Two convex sets in $\mathbb{R}^n$. Only give practical approximate formulae for balls, boxes and cylinders.
        & No
        & No, with arbitrarily small error as $h\to+0$
        & Iterative, no fixed number of steps \\
        \hline

        Beker et al. (2025) \cite{beker2025smooth}
        & Almost arbitrary sets in $\mathbb{R}^n$,
          represented as point clouds
        & Yes
        & No, with arbitrarily small error as $h\to+0$
        & $\mathcal{O}(K_{\mathcal{A}}K_{\mathcal{B}})$,
          where $K_{\mathcal{A}}$ and $K_{\mathcal{B}}$
          are the numbers of points in the respective objects \\
        \hline

        Wu et al. (2025) \cite{11060906}
        & Two convex polyhedral sets in $\mathbb{R}^2$
        & Yes
        & No; conservative for finite smoothing,
          with arbitrarily small error as $h\to +0$
        & $\mathcal{O}(V_{\mathcal{A}}V_{\mathcal{B}})$,
          where $V_{\mathcal{A}}$ and $V_{\mathcal{B}}$
          are the numbers of vertices \\
        \hline

        Gonçalves et al. (2026) \cite{Goncalves2026}
        & Two convex sets in $\mathbb{R}^n$. Only gives practical formulae for polyhedra.
        & No
        & Yes
        & Iterative, no fixed number of steps. If the two objects are polyhedral and have $F_{\mathcal{A}}, F_{\mathcal{B}}$ faces, respectively, each step is $\mathcal{O}(F_{\mathcal{A}}+F_{\mathcal{B}})$  \\
        \hline

        Nunes et al. (2026) \cite{Nunes2026}
        & Point cloud and a point in $\mathbb{R}^n$
        & No
        & Yes
        & $\mathcal{O}(K)$,
          where $K$ is the number of points \\
        \hline

        Bartelt et al. (2026) \cite{bartelt2026h}
        & Two convex polyhedral sets in $\mathbb{R}^3$
        & Yes
        & Not exactly, but collision detection can become
          arbitrarily accurate by increasing the number
          of normals $N$
        & $\mathcal{O}(V_{\mathcal{A}}
          V_{\mathcal{B}}N)$,
          where $V_{\mathcal{A}}$ and $V_{\mathcal{B}}$
          are the numbers of vertices and $N$ is
          the number of normals used. The paper proposes, by default, $N = \mathcal{O}(V_{\mathcal{A}}+V_{\mathcal{B}})$. \\
        \hline

        Ours
        & Almost arbitrary sets in $\mathbb{R}^n$, one of them at least
          represented as a point cloud
        & Yes
        & No, with arbitrarily small error as $h\to+0$
        & If $\mathcal{A}$ is represented by a point cloud with $K_{\mathcal{A}}$ points, the complexity is $\mathcal{O}(K_{\mathcal{A}})$  
          calls of a function related to $\mathcal{B}$. If $\mathcal{B}$ is also a point cloud, each call is $\mathcal{O}(K_{\mathcal{B}})$,  resulting in $\mathcal{O}(K_{\mathcal{A}}K_{\mathcal{B}})$. \\
        \hline

    \end{tabularx}
\end{table*}

In this work, we propose a signed, infinitely differentiable distance-like function between two possibly nonconvex objects. We first introduce the \emph{Normal Euclidean Signed Distance Function} (NESDF), a potentially nonsmooth function whose sign characterizes separation, contact, and interior overlap. We then apply a smoothing procedure to obtain the $h$-NESDF, which is infinitely differentiable with respect to smoothly parameterized rigid motions. Its sampled approximation is amenable to GPU parallelization. Figure \ref{fig:nesdf_comparison} shows an example of the $h$-NESDF computed for two $L$ shapes as one of them rotates, with three different values of $h$, one of them being as $h \rightarrow +0$, that is, the non-differentiable case.

Our construction differs from the proposed metric in \cite{beker2025smooth} in several respects, bringing some benefits. One distinction concerns the use of the gradient of the metric in a gradient-ascent procedure for resolving overlap between two objects, useful, for example, in motion planning. In some deeply overlapping configurations, the metric in \cite{beker2025smooth} produces an ascent direction that increases geometric overlap, whereas ours produces the correct behavior.  These distinctions are examined in detail throughout the paper. 

The \textbf{contributions} of this paper are then:

\begin{enumerate}
    \item NESDF, the \emph{Normal Euclidean Signed  Distance Function}, a novel signed distance metric between two general, possibly non-convex sets in $\mathbb{R}^n$. We establish that this metric has the expected behavior of being positive when the two objects do not overlap and negative when they overlap. This metric is also amenable for a smoothfication procedure.
    \item $h$-NESDF, a smoothed version of NESDF that is infinitely differentiable, with smoothing parameter $h$ that controls the smoothing. We show that when $h \rightarrow +0$, NESDF is recovered. This metric can be implemented in an approximated manner using sampled point clouds and normal vectors, and parallelized on a GPU. This metric brings benefits when compared to other comparable approaches, as \cite{beker2025smooth}.
    \item We present extensive computational experiments assessing $h$-NESDF efficiency and usefulness in motion planning .
\end{enumerate}

For readability, all proofs and auxiliary lemmas are deferred to the Appendix.

\subsection{Basic Notation}

Any vector is a column vector unless stated otherwise. $A^{\top}$ represents the transpose of a matrix. For $f: \mathbb{R}^n \mapsto \mathbb{R}$, $\nabla f$, the gradient, is a column vector. For a set $\mathcal{X}$, $\partial \mathcal{X}$ represents its boundary, $\textsl{int}(\mathcal{X})$ its interior, and $\overline{\mathcal{X}}$ its closure. $\mathbb{S}_n$ represents the set of $n$-dimensional normalized vectors. $\mathcal{X}-\mathcal{Y}$ represents the set subtraction (all points in $\mathcal{X}$ not in $\mathcal{Y}$). Uniform convergence of $f_h(y)$ to $f(y)$ as $h \rightarrow +0$ in a set $\mathcal{Y}$ means that for any $\delta>0$, there exists $\epsilon(\delta, \mathcal{Y})$ such that

\begin{equation}
|h| < \epsilon \ \rightarrow \sup_{y \in \mathcal{Y}} \|f_h(y)-f(y)\| < \delta.
\end{equation}

\section{Methodology}

\subsection{The Normal Euclidean Signed  Distance Function (NESDF)}

In this Subsection, we will propose NESDF, the \emph{ Normal Euclidean Signed  Distance Function}, a novel signed metric between two nonconvex shapes. It will be used as a starting point for the true goal of this work, which is to define an infinitely differentiable metric between two objects.  We start with the following definition.

\begin{definition}\emph{(Euclidean Signed Distance Function (ESDF) between a point and a set)} Let $y \in \mathbb{R}^n$ and let $\mathcal{X}$ be a compact subset of $\mathbb{R}^n$. The ESDF $D_{\mathcal{X}}(y)$ between $y$ and $\mathcal{X}$ is defined as follows \cite{OsherFedkiw2003}:
\begin{equation}
  D_{\mathcal{X}}(y)
\triangleq
\begin{cases}
\ \displaystyle \ \ \min_{x \in \mathcal{X}} \|x-y\|, & y \not \in  \mathcal{X},\\
- \displaystyle \min_{x \in \partial \mathcal{X}} \|x-y\|, & \mbox{otherwise}.
\end{cases}
\end{equation}
$\blacksquare$
\end{definition}

There is no single standard definition of the ESDF between two \emph{sets} $\mathcal{A}$ and $\mathcal{B}$. One definition uses penetration depth \cite{SchulmanEtAl2014}; alternatively, one can minimize the point-to-set ESDF over the other set and define $D_{\mathcal{A},\mathcal{B}} = \min_{b \in \mathcal{B}} D_{\mathcal{A}}(b)$, as done for point-cloud clearance in \cite{KuntzFuAlterovitz2019}. Neither is readily amenable to smoothing. We therefore will introduce NESDF, that is easier to smooth. First, we need some auxiliary definitions.

\begin{definition}
    \emph{(Regular Set)} In this paper, a set $\mathcal{X} \subseteq \mathbb{R}^n$ is called \emph{regular} if:
    \begin{enumerate}
        \item It is compact (i.e., closed and bounded);
        \item Its interior is nonempty and $\mathcal{X}=\overline{\operatorname{int}(\mathcal{X})}$;
        \item Its boundary is orientable;
        \item The normal vector at the boundary exists almost everywhere (i.e., the set of points at which it does not exist has zero measure in $\mathbb{R}^n$).
        \item Its interior is path connected: any two points $x_1$ and $x_2$ in the interior of $\mathcal{X}$ can be connected by a continuous path in its interior.
    \end{enumerate}
$\blacksquare$
\end{definition}

\begin{definition}
    \emph{(Normal Function)} Let $\mathcal{X}$ be a regular set. For any point $x \in \mathcal{X}$ such that the normal vector at the boundary is defined, let $n_{\mathcal{X}}: \partial \mathcal{X} \mapsto \mathbb{S}^{n-1}$ as the outward-pointing normal. $\blacksquare$
\end{definition}

\begin{definition}
    \emph{(Half-Squared ESDF)} Let $\mathcal{X}$ be a regular set. Let $d_{\mathcal{X}} \triangleq \min_{x} D_{\mathcal{X}}(x)$ which is finite if $\mathcal{X}$ is compact. Define $E_{\mathcal{X}}(y) \triangleq 0.5(D_{\mathcal{X}}(y)-d_{\mathcal{X}})^2-0.5d_{\mathcal{X}}^2$. $\blacksquare$
\end{definition}

The function $E_{\mathcal{X}}(y)$ generalizes the half-squared Euclidean distance function used in the unsigned case, where $d_{\mathcal{X}}=0$ (e.g., \cite{Goncalves2024,Goncalves2026}) to account for signed distances.

We can now define, under some circumstances, the \emph{ Normal Euclidean Signed  Distance Function} (NESDF). 

\begin{definition} \emph{(Non-Degenerate Pair and NESDF)}
 We say that a pair of sets $(\mathcal{A},\mathcal{B})$  is a \emph{nondegenerate pair} if

\begin{enumerate}
\item  $\mathcal{A}$ and $\mathcal{B}$ are regular sets;
    \item The optimizer 
    \begin{equation}
    \label{eq:bstar}
    b^* \triangleq \arg \min_{b \in \partial \mathcal{B}} E_{\mathcal{A}}(b)
\end{equation}

\noindent is unique;
\item $E_{\mathcal{A}}(b)$ is differentiable in a neighborhood of $b = b^*$;
\item The normal vector $n_{\mathcal{B}}(b)$ is continuous in a neighborhood of $b=b^*$.
\end{enumerate}
Then, for a nondegenerate pair, define the NESDF as

\begin{equation}
    N_{\mathcal{A},\mathcal{B}} \triangleq d_{\mathcal{A}}-n_{\mathcal{B}}(b^*)^{\top} \nabla E_{\mathcal{A}}(b^*).
\end{equation}
$\blacksquare$
\end{definition}

The name reflects the role of the normal vector $n_{\mathcal{B}}$. At this stage, the definition of the NESDF is incomplete, since it is not defined for pairs that are not nondegenerate. This is not a problem: the NESDF was only created to motivate its smoothed version, in which this nondegeneracy hypothesis will no longer be necessary.

We will now show some properties of the NESDF, starting with the most important one.

\begin{proposition} \label{eq:signresult}
Let $(\mathcal{A},\mathcal{B})$ be a nondegenerate pair. Then:
\begin{enumerate}
    \item \textbf{Case 1}: When  $\textsl{int}(\mathcal{A}) \cap \textsl{int}(\mathcal{B}) = \emptyset$ and    $\partial \mathcal{A} \cap \partial \mathcal{B} = \emptyset$, $N_{\mathcal{A},\mathcal{B}} > 0$;
    \item \textbf{Case 2}: When  $\textsl{int}(\mathcal{A}) \cap \textsl{int}(\mathcal{B}) = \emptyset$ and    $\partial \mathcal{A} \cap \partial \mathcal{B} \not= \emptyset$, $N_{\mathcal{A},\mathcal{B}} = 0$;
    \item \textbf{Case 3}:  When  $\textsl{int}(\mathcal{A}) \cap \textsl{int}(\mathcal{B}) \not= \emptyset$,   $N_{\mathcal{A},\mathcal{B}} < 0$.
\end{enumerate}
$\blacksquare$
\end{proposition}

Thus, the NESDF has the expected sign behavior of a signed distance between two objects. In addition, in one particular case, the NESDF is exactly the ESDF.

\begin{proposition}
    If the formula for $N_{\mathcal{A},\mathcal{B}}$ is extended to a singleton $\mathcal{A} = \{a\}$, then, if $(\{a\},\mathcal{B})$ is a non-degenerate pair, $N_{\mathcal{A},\mathcal{B}} = D_{\mathcal{B}}(a)$. $\blacksquare$
\end{proposition}

\subsection{The $h$-NESDF}

In this subsection, we will show how to use the NESDF to obtain a smoothed, infinitely differentiable version of this metric. First, we assume a smoothed version of the ESDF:

\begin{definition}\label{def:hesdf}
\emph{($h$-ESDF)}
Let $\mathcal X\subseteq\mathbb R^n$ be a regular set.
Consider a family $D_{\mathcal X}^h:\mathbb R^n\to\mathbb R$,
parameterized by $h>0$, and define
\begin{equation}
d_{\mathcal X}^h\triangleq
\min_{y\in\mathbb R^n}D_{\mathcal X}^h(y),
\end{equation}
\begin{equation}
\label{eq:eXh}
E_{\mathcal X}^h(y)\triangleq
\frac12\bigl(D_{\mathcal X}^h(y)-d_{\mathcal X}^h\bigr)^2
-\frac12(d_{\mathcal X}^h)^2.
\end{equation}
We call this family an $h$-ESDF if:
\begin{enumerate}
\item $E_{\mathcal X}^h$ is infinitely differentiable for any $h>0$.
\item $d_{\mathcal X}^h\to d_{\mathcal X}$ when $h \to +0$;
\item $E_{\mathcal X}^h\to E_{\mathcal X}$ when $h \to +0$ uniformly in every compact set;
\item $\nabla E_{\mathcal X}^h\to\nabla E_{\mathcal X}$
as $h\to0^+$, uniformly on every compact set contained
in an open set on which $E_{\mathcal X}$ is
continuously differentiable;
\item For any $\Delta > 0$, $e^{-\Delta/h^2}\nabla E_{\mathcal X}^h \to 0$ when $h \to +0$
uniformly on every compact set.
\end{enumerate}
$\blacksquare$
\end{definition}

The last four properties just guarantee that $E^h_{\mathcal X} \to E_{\mathcal X}$ with sufficient regularity for the convergence result below. See Subsection \ref{subs:compute_D} for examples on how to compute the $h$-ESDF. Nevertheless, using $D_{\mathcal{A}}^h(b)$, we can define the smoothed version of our NESDF:

\begin{definition}
    \emph{($h$-NESDF)} Consider two regular sets $\mathcal{A}$, $\mathcal{B}$ and a $h$-ESDF $D_{\mathcal{A}}^h$. Define a random variable $B_{\mathcal{A},h}$, parametrized by the set $\mathcal{A}$ and $h>0$, with values in $\partial \mathcal{B}$ and density

\begin{equation}
\label{eq:rho}
    \rho_{\mathcal{A},h}(b) = \frac{e^{-E_{\mathcal{A}}^h(b)/h^2}}{\int_{\partial \mathcal{B}} e^{-E_{\mathcal{A}}^h(\beta)/h^2}dS}
\end{equation}

\noindent (in which $E_{\mathcal{A}}^h$ was defined in \eqref{eq:eXh}) for $b \in \partial \mathcal{B}$, where $dS$ is the surface element on $\partial\mathcal{B}$. We can then define the $h$-NESDF as

\begin{equation}
   N_{\mathcal{A},\mathcal{B}}^h \triangleq d_{\mathcal{A}}^h -\mathbb{E}\Big[n_{\mathcal{B}}\big(B_{\mathcal{A},h}\big)^{\top}\nabla E_{\mathcal{A}}^h\big(B_{\mathcal{A},h}\big)\Big]. 
\end{equation}

Or, more explicitly:

\begin{equation}
   N_{\mathcal{A},\mathcal{B}}^h \triangleq d_{\mathcal{A}}^h-\frac{\int_{\partial \mathcal{B}} e^{-E_{\mathcal{A}}^h(\beta)/h^2}n_{\mathcal{B}}(\beta)^{\top}\nabla E_{\mathcal{A}}^h(\beta)dS}{\int_{\partial \mathcal{B}} e^{-E_{\mathcal{A}}^h(\beta)/h^2}dS}.
\end{equation}
$\blacksquare$
\end{definition}

Note that, as opposed to the definition of NESDF, the $h$-NESDF does not require that $(\mathcal{A},\mathcal{B})$ is a nondegenerate pair. In addition, note that, since $\mathcal{B}$ is assumed regular, the set of points $b$ in which $n_{\mathcal{B}}(b)$ is undefined has zero measure in $\mathbb{R}^n$, and thus their choice of values is irrelevant for the value of the integral that defines $N_{\mathcal{A},\mathcal{B}}^h$.

\begin{figure}[!t]
    \centering
    \includegraphics[width=0.7\columnwidth]{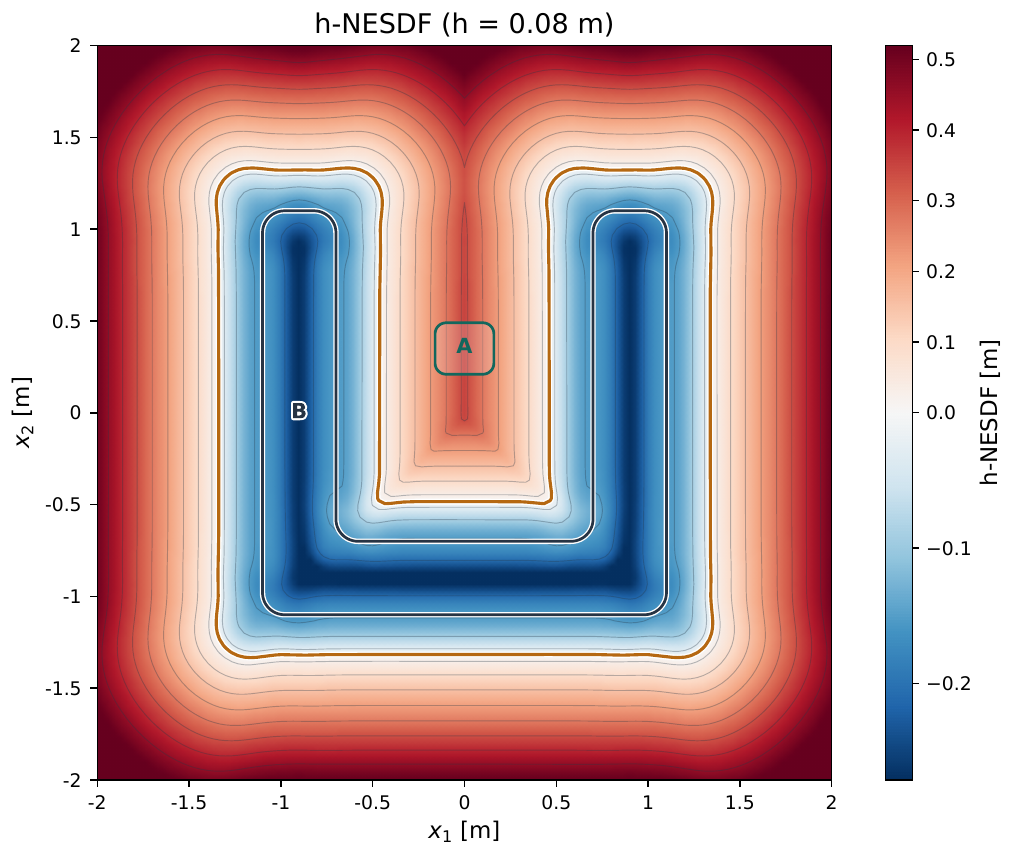}
    \caption{Contour curves of the $h$-NESDF ($h=0.08\,\mathrm{m}$) for a nonconvex U-shaped object $\mathcal{B}$ and a rounded square object $\mathcal{A}$ as the center of $\mathcal{A}$ moves. The black thick contour represents the boundary of $\mathcal{B}$, whereas the outer brown contour indicates where the \(h\)-NESDF is zero }
    \label{fig:hnesdf_field_2d}
\end{figure}

Figure \ref{fig:hnesdf_field_2d} shows, in the 2D case, the h-NESDF between two objects, a rounded square $\mathcal{A}$ and an $U$-shaped object $\mathcal{B}$. More precisely, it shows the contour curves of the function $f: \mathbb{R}^2 \mapsto \mathbb{R}$ defined by $f(s) = N_{\mathcal{A}(s),\mathcal{B}}^h$, in which $\mathcal{A}(s)$ represents the rounded square depicted in Figure \ref{fig:hnesdf_field_2d}  with center $s \in \mathbb{R}^2$.

As $h \rightarrow 0^+$, under some conditions, $N_{\mathcal{A},\mathcal{B}}^h \rightarrow N_{\mathcal{A},\mathcal{B}}$.

\begin{proposition} Let  $(\mathcal{A},\mathcal{B})$ be a nondegenerate pair. Then:
\begin{equation}
    \lim_{h \rightarrow +0} N_{\mathcal{A},\mathcal{B}}^h = N_{\mathcal{A},\mathcal{B}}.
\end{equation}
$\blacksquare$
\end{proposition}

Note that the $h$-NESDF  does not have exactly the same sign properties as the NESDF shown in Proposition \ref{eq:signresult}. Thus, it is not guaranteed to be positive when the objects do not touch or negative when their interiors overlap. However, the $h$-NESDF implements an approximate version of this result, approximating the original result as $h \rightarrow +0$ at the expense of potentially larger derivative magnitudes and sharper transitions. Even though the exact sign property in Proposition \ref{eq:signresult} does not hold, it is still useful in robotic control and motion planning with the appropriate care, as will be shown.

We will now state results about the differentiability. We will assume that the objects $\mathcal{A}(t) \subseteq \mathbb{R}^n$ and $\mathcal{B}(t) \subseteq \mathbb{R}^n$ move as \emph{rigid bodies}. Thus, there exists $Q_{\mathcal{A}}(t), Q_{\mathcal{B}}(t) \in \textsl{SO}(n)$ and $s_{\mathcal{A}}(t), s_{\mathcal{B}}(t) \in \mathbb{R}^n$ along with static sets $\mathscr{a}, \mathscr{b}$ such that $\mathcal{A}(t) = Q_{\mathcal{A}}(t)\mathscr{a}+s_{\mathcal{A}}(t)$ and  $\mathcal{B}(t) = Q_{\mathcal{B}}(t)\mathscr{b}+s_{\mathcal{B}}(t)$. Given the $h$-ESDF $E^h_{\mathscr{a}}(b)$, we use:

\begin{equation}
\label{eq:Dh_transf}
    D^h_{\mathcal{A}(t)}(b) = D^h_{\mathscr{a}}\big(Q_{\mathcal{A}}(t)^{\top}(b-s_{\mathcal{A}}(t))\big).
\end{equation}

This choice guarantees that

\begin{equation}
d^h_{\mathcal{A}(t)} = \min_{a} D^h_{\mathcal{A}(t)}(a) = \min_{a} D^h_{\mathscr{a}}(a) = d^h_{\mathscr{a}}
\end{equation}

\noindent is \emph{constant}, avoiding a possible source of non-differentiability due to the fact that minimum is not a differentiable operator. Then:

\begin{proposition}
    If $Q_{\mathcal{A}}(t), Q_{\mathcal{B}}(t)$ and $s_{\mathcal{A}}(t), s_{\mathcal{B}}(t)$ are $k$-times differentiable in $t$, so is 
    $f(t) = N^h_{\mathcal{A}(t),\mathcal{B}(t)}$. $\blacksquare$ 
\end{proposition}

Thus, when the motions are infinitely differentiable, so is the distance function.

\section{Implementation}

\subsection{Implementing the $h$-NESDF in practice}
\label{subs:implpractice}

Assuming that we have the $h$-ESDF function $D_{\mathcal{A}}^h(b)$ (see the next subsection), we sample $K$ points $b_k$ on $\partial\mathcal{B}$ with respective normals $n_k$, obtaining $K$ pairs $(b_k,n_k)$. Since the formula for $N_{\mathcal{A},\mathcal{B}}^h$ is a weighted average, we replace the integrals by sums, assuming that the points $b_k$ are distributed approximately uniformly over the surface:

\begin{equation}
\label{eq:NABpoints}
   N_{\mathcal{A},\mathcal{B}}^h  \approx  d_{\mathcal{A}}^h -\frac{\sum_{k=1}^K e^{-E_{\mathcal{A}}^h(b_k)/h^2}n_k^{\top}\nabla E_{\mathcal{A}}^h(b_k)}{\sum_{k=1}^K e^{-E_{\mathcal{A}}^h(b_k)/h^2}} 
\end{equation}

For numerical stability reasons, one should subtract from each $E_{\mathcal{A}}^h(b_k)$ the value $\delta = \min_k E_{\mathcal{A}}^h(b_k)$. This does not modify the result theoretically, but in practice help avoiding underflow with the exponentials.

\subsection{Computing $h$-ESDF for arbitrary $\mathcal{X}$}
\label{subs:compute_D}

\begin{figure*}[!t]
    \centering
    \includegraphics[width=1\textwidth]{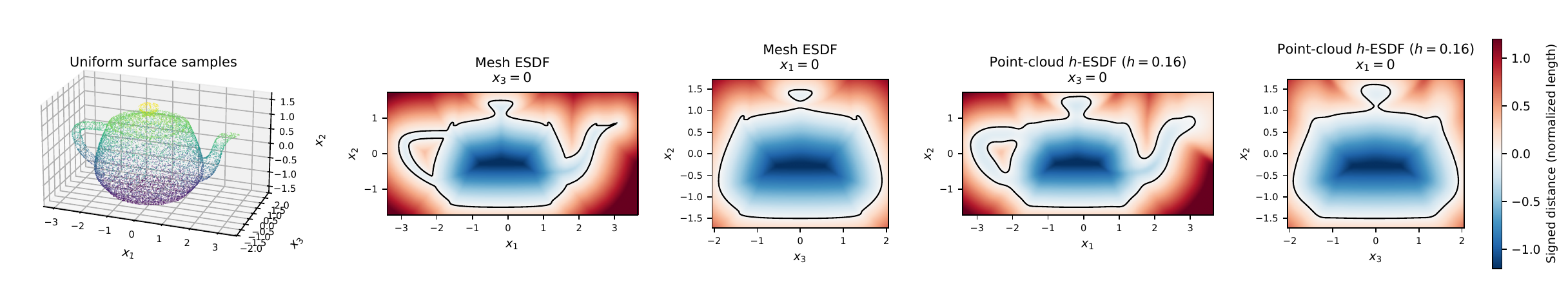}
    \caption{The Utah-Teapot, sampled as points and normals, and the ESDF (computed using the mesh) and $h$-ESDF (computed using sampled points and normals) signed distance fields. The thick black outline shows the zero level sets. This was computed with the approach described in Subsection \ref{subs:compute_D}.}
    \label{fig:five-panel-esdf-pc}
\end{figure*}

\begin{figure*}[!t]
    \centering
    \includegraphics[width=0.9\textwidth]{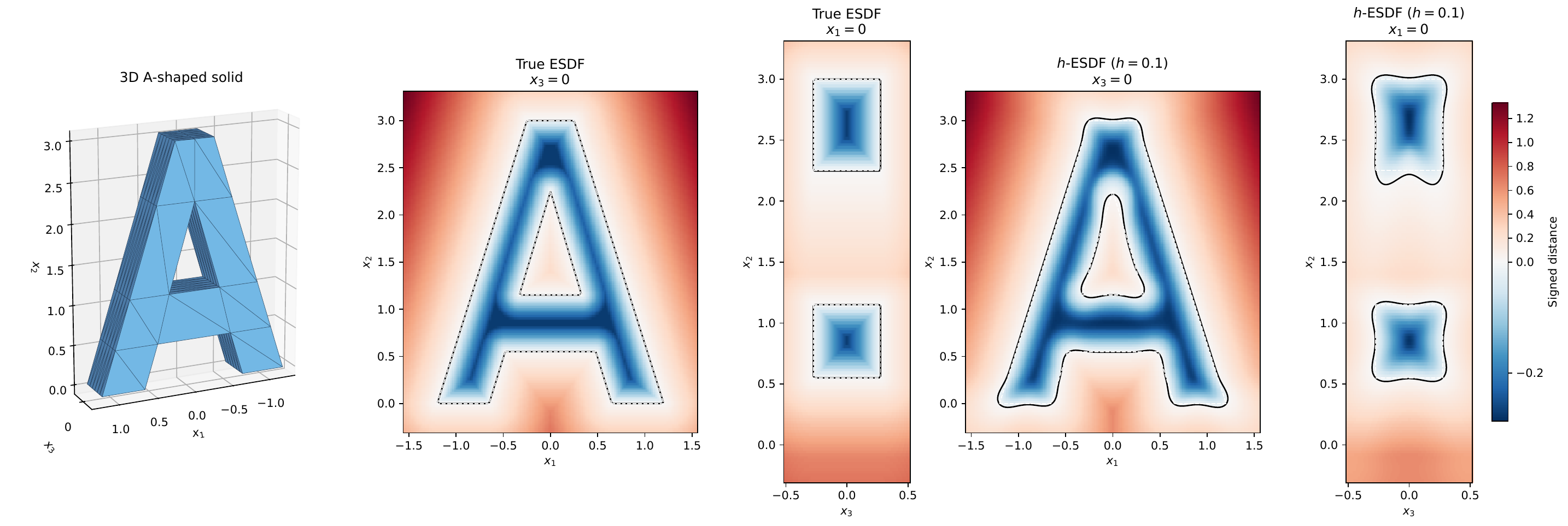}
    \caption{An A-shaped solid and slices of the ESDF and $h$-ESDF. The thick black outline shows the zero level sets.  This was computed with the approach described in Subsection \ref{subs:compute_D} for polyhedra.}
    \label{fig:five-panel-esdf}
\end{figure*}

To obtain an $h$-ESDF $D_{\mathcal X}^h$ for a regular set $\mathcal X$,
we can use the construction in Subsection \ref{subs:implpractice} with
$\mathcal A=\{y\}$ and $\mathcal B=\mathcal X$. In this case, we can take
$D_{\mathcal A}^h(b)=D_{\mathcal A}(b) = \|b-y\|$ for every $h$, since $E_{\mathcal{A}}^h(b) = E_{\mathcal{A}}(b) = 0.5\|b-y\|^2$ is infinitely differentiable.
We therefore obtain
\begin{equation}
\label{eq:formulaEX}
D_{\mathcal X}^h(y)=
\frac{
\int_{\partial\mathcal X}
e^{-\|x-y\|^2/(2h^2)}
n_{\mathcal X}(x)^\top(y-x)\,dS
}{
\int_{\partial\mathcal X}
e^{-\|x-y\|^2/(2h^2)}\,dS
}.
\end{equation}

This formula can be applied to different representations of
$\mathcal X$. For example, when its boundary is represented by
sampled point--normal pairs, we can use the approximation discussed
in Subsection \ref{subs:implpractice}. See Figure \ref{fig:five-panel-esdf-pc} for an example of the $h$-ESDF implemented with this technique. In this case, given a set of points and normals $(x_k,n_k)$ sampled approximately uniformly on $\partial \mathcal{X}$, the formula reduces to 
\begin{equation}
\label{eq:formulaEXapprox}
  D_{\mathcal X}^h(y)\approx
\frac{
\sum_{k=1}^K
e^{-\|x_k-y\|^2/(2h^2)}
n_{k}^\top(y-x_k)
}{
\sum_{k=1}^Ke^{-\|x_k-y\|^2/(2h^2)}}.  
\end{equation}

\noindent which is essentially a result that appeared in other papers, as  in \cite{beker2025smooth, 9578427}.

Now suppose that $\mathcal X\subset\mathbb R^n$ is a possibly
non-convex polyhedron. Partition its boundary into $K$
$(n-1)$-dimensional polyhedral pieces $\mathcal X_k$.
Each piece lies in a plane $n_k^\top x+c_k=0$, where $n_k$
is a unit outward normal. Define
\begin{equation}
\omega_k^h(y)=
\int_{\mathcal X_k}e^{-\|x-y\|^2/(2h^2)}\,dS.
\end{equation}
Equation \eqref{eq:formulaEX} then becomes
\begin{equation}
\label{eq:Eformula_polyhedra}
D_{\mathcal X}^h(y)=
\frac{\sum_{k=1}^K\omega_k^h(y)(n_k^\top y+c_k)}
{\sum_{k=1}^K\omega_k^h(y)}.
\end{equation}

All weights in \eqref{eq:Eformula_polyhedra} can underflow when
$y$ is far from the boundary. To alleviate this,
we can multiply every weight by the same factor
$e^{\delta(y)/h^2}$, for any function $\delta(y)$, simiarly as it was done in Subsection \ref{subs:implpractice}. We then obtain
\begin{equation}
\label{eq:shifted_weights}
\widetilde\omega_k^h(y)=
\int_{\mathcal X_k}
e^{
(\delta(y)-0.5\|x-y\|^2)/h^2
}dS.
\end{equation}
Since
$\omega_k^h(y)=e^{-\delta(y)/h^2}
\widetilde\omega_k^h(y)$, the common factor cancels:
\begin{equation}
\label{eq:shifted_polyhedra}
D_{\mathcal X}^h(y)=
\frac{\sum_{k=1}^K\widetilde\omega_k^h(y)
(n_k^\top y+c_k)}
{\sum_{k=1}^K\widetilde\omega_k^h(y)}.
\end{equation}

The ideal $\delta$ is $\delta(y) = 0.5 D_{\partial \mathcal{X}}(y)^2$, but it can be relatively expensive to compute. We can use a relatively inexpensive proxy. Let $\{v_i\}$ contain the vertices
of all the pieces $\mathcal{X}_k$. Let
\begin{equation}
\delta(y)=0.5\min_i\|y-v_i\|^2.
\end{equation}

In three dimensions, we can take the pieces $\mathcal X_k$ to
be triangles. Let $\mathcal S(a,b,c)$ be a canonical 2D triangle
with vertices $(0,0)$, $(a,0)$, and $(b,c)$, where
$a,c>0$. Let $Q_k \in\mathrm{SO}(3), s_k \in \mathbb{R}^3$ be the rigid transformation that maps the 2D canonical triangle, after lifting it to $3D$ by adding a $0$ coordinate, to $\mathcal{X}_k$. For $z=[z_1 \ z_2 \ z_3]^\top$ and $\zeta = [\zeta_1 \ \zeta_2]^{\top}$, define
\begin{eqnarray}
    &&\Delta(\zeta;z,\delta) \triangleq  \delta{-}0.5(\zeta_1{-}z_1)^2{-}0.5(\zeta_2{-}z_2)^2-0.5z_3^2.  \nonumber \\   
    &&\Phi^h(z;a,b,c,\delta) \triangleq \int_{\mathcal{S}(a,b,c)} e^{\Delta(\zeta;z,\delta)/h^2}d\zeta_1 d\zeta_2.
\end{eqnarray}

Then:
\begin{equation}
    \tilde{\omega}_k^h(y) = \Phi^h\big(Q_k^{\top}(y-s_k);a_k,b_k,c_k,\delta(y)\big).
\end{equation}
This function $\Phi^h$ can be approximated using Gauss-Legendre quadrature in 2D after using Duffy Transformation \cite{Badia2018FEMPAR} to map the triangular region to a square. The nodes and weights should be selected independent of $y$, so the resulting approximation continues infinitely differentiable in $y$. Figure \ref{fig:five-panel-esdf} shows the result when this polyhedral procedure is implemented for an ``A'' shaped polyhedra.




\section{Experimental Evaluation and Comparisons}

\subsection{Qualitative Comparison with \cite{beker2025smooth}}
\label{subs:qualcomp}

As mentioned in the introduction, \cite{beker2025smooth} also proposes a differentiable distance metric between two general sets. Just like in our proposed approach, they start from a non-differentiable counterpart and then apply a smoothing procedure.

In their case, they start from:

\begin{equation}
    M_{\mathcal{A},\mathcal{B}} = \min\Big(\min_{a \in \partial \mathcal{A}} D_{\mathcal{B}}(a), \min_{b \in \partial \mathcal{B}} D_{\mathcal{A}}(b) \Big)
\end{equation}

\noindent and then smooth it with a parameter $h$, obtaining a metric  $M_{\mathcal{A},\mathcal{B}}^h$. Their procedure to smooth this formula is similar to what is proposed here, and includes a smoothfication of $D_{\mathcal{X}}(y)$ into $D^h_{\mathcal{X}}(y)$ that is essentially a slight generalization of \eqref{eq:formulaEXapprox} (just with a general function $B_h(x-y)$ instead of $e^{-\|x-y\|^2/(2h^2)}$.  Nevertheless, following the implementation in \cite{beker2025smooth}, we will consider, if $\mathcal{A}$ and $\mathcal{B}$ has $K_{\mathcal{A}}$ and $K_{\mathcal{B}}$ points respectively, we will define $M^h_{\mathcal{A},\mathcal{B}}$ as

\begin{equation}
  \approx \frac{\sum_{k=1}^{K_{\mathcal{A}}} e^{-D^h_{\mathcal{B}}(a_k)/h}D^h_{\mathcal{B}}(a_k) + \sum_{k=1}^{K_{\mathcal{B}}} e^{-D^h_{\mathcal{A}}(b_k)/h}D^h_{\mathcal{A}}(b_k) }{\sum_{k=1}^{K_{\mathcal{A}}} e^{-D^h_{\mathcal{B}}(a_k)/h} + \sum_{k=1}^{K_{\mathcal{B}}}e^{-D^h_{\mathcal{A}}(b_k)/h}}
\end{equation}

\noindent which is comparable with the point-cloud based formula that we propose in \eqref{eq:NABpoints}. 

Since both approaches start with a non-differentiable metric and smooth it out, obtaining approximations of their original counterparts, it is fair to begin by comparing their unsmoothed counterparts: $N_{\mathcal{A},\mathcal{B}}$ and $M_{\mathcal{A},\mathcal{B}}$.

Let $\mathcal{A}$ be a sphere with center $q \not=0$ and radius $R$, and $\mathcal{B}$ a sphere with center $0$ and radius $R$. One can clearly see that $D_{\mathcal{A}}(b) = \|b-q\|-R \ , \ D_{\mathcal{B}}(a) = \|a\|-R$ and also that:

\begin{equation}
    \min_{a \in \partial \mathcal{A}} D_{\mathcal{B}}(a) = \min_{b \in \partial \mathcal{B}} D_{\mathcal{A}}(b) =  |\|q\|-R|-R
\end{equation}
\noindent and thus $M_{\mathcal{A},\mathcal{B}} = |\|q\|-R|-R$.

Now, $b^* = \arg \min_{b \in \partial \mathcal{B}} E_{\mathcal{A}}(b) = Rq/\|q\|$. In this case, $n_{\mathcal{B}}(b^*) = q/\|q\|$, $d_{\mathcal{A}} = -R$ and $\nabla E_{\mathcal{A}}(b^*) = (b^*-q) = Rq/\|q\|-q$. Thus $N_{\mathcal{A},\mathcal{B}} = -R-(R-\|q\|) = \|q\|-2R$.

The difference now is clear, $M_{\mathcal{A},\mathcal{B}}(q) = |\|q\|-R|-R$ and $N_{\mathcal{A},\mathcal{B}}(q) =  \|q\|-2R$. If one uses the gradient of these metrics on $q$ to move $q$ to make the sphere no longer overlap, $N_{\mathcal{A},\mathcal{B}}$ gives the proper direction $\nabla N_{\mathcal{A},\mathcal{B}}(q) = q/\|q\|$, whereas $\nabla M_{\mathcal{A},\mathcal{B}}(q) = \textsl{sign}(\|q\|-R)q/\|q\|$. Thus, if $\|q\|<R$, then $\nabla M_{\mathcal{A},\mathcal{B}}(q) = -q/\|q\|$, making the two spheres overlap even more.

\subsection{Comparison with \cite{beker2025smooth}: Escape from Overlapping Configurations}
\label{subsec:escape_comparison}

The example in the previous subsection shows that the gradient of a signed distance-like function may not always provide a suitable direction for resolving interpenetration. We therefore compare the $h$-NESDF with the smooth metric of \cite{beker2025smooth}, named hereafter ``Beker'' in a large-scale rigid-body separation task. Two initially overlapping objects, $\mathcal A$ and $\mathcal B$, are considered: $\mathcal A$ remains fixed, while $\mathcal B$ can translate and rotate freely in $\textsl{SE}(3)$. For each method, the objective $f$ is its corresponding smoothed metric, $N^h_{\mathcal A,\mathcal B}$ or $M^h_{\mathcal A,\mathcal B}$, we maximize the corresponding smoothed function, evaluated at the current relative pose.

Let $f(Q_{\mathcal A},s_{\mathcal A},Q_{\mathcal B},s_{\mathcal B})$
denote either the proposed $h$-NESDF or the smooth metric of
\cite{beker2025smooth}, expressed as a function of the rigid-body
poses. We consider
the row vector $L[f](Q_{\mathcal A},s_{\mathcal A},Q_{\mathcal B},s_{\mathcal B}) \in\mathbb R^{1 \times 12}$, defined such that
\begin{equation}
    \label{eq:escape_operator_full}
    \frac{d}{dt} f(Q_{\mathcal A},s_{\mathcal A},Q_{\mathcal B},s_{\mathcal B}) = L[f](Q_{\mathcal A},s_{\mathcal A},Q_{\mathcal B},s_{\mathcal B})\xi,
\end{equation} 

 \noindent where $\xi \in \mathbb{R}^{12}$ is a column vector given by 

\begin{equation}
    \xi = [v_{\mathcal{A}}^{\top} \ \omega_{\mathcal{A}}^{\top} \ v_{\mathcal{B}}^{\top} \ \omega_{\mathcal{B}}^{\top}]^{\top}.
\end{equation}
Here, $\omega_i$ and $v_i$ denote, respectively, the angular
velocity and the linear velocity vectors of each body, both expressed
in world coordinates.

The operator $L[f]$ plays the role of the gradient $\nabla f$ if its arguments were Euclidean vectors.  The row vector $L[f]$ can be
partitioned as
$L[f]=[L_{\mathcal A}[f] \
L_{\mathcal B}[f]]$, corresponding to the two bodies, respectively.

In the separation experiments, $\mathcal A$ remains fixed,
while $\mathcal B$ is free to move in $\textsl{SE}(3)$.
Consequently, $\xi_{\mathcal A}=0$ and
\begin{equation}
    \label{eq:escape_operator}
    \dot f =
    L_{\mathcal B}[f](Q_{\mathcal A},s_{\mathcal A},Q_{\mathcal B},s_{\mathcal B})\xi_{\mathcal B}.
\end{equation}
Therefore, at a given configuration $(Q_{\mathcal A},s_{\mathcal A},Q_{\mathcal B},s_{\mathcal B})$ of both objects, $L_{\mathcal B}[f](Q_{\mathcal A},s_{\mathcal A},Q_{\mathcal B},s_{\mathcal B})$ provides the local
ascent direction of the selected metric with respect to
the six degrees of freedom of $\mathcal B$.

To perform the separation task, we employ a gradient-ascent, with $L_{\mathcal{B}}[f]$ playing the role of the gradient, controller with Nesterov acceleration \cite{nesterov1983method}. Let $S: \mathbb{R}^6 \mapsto \textsl{se}(3)$ be the linear operator such that, for any $p \in \mathbb{R}^3$, $q \in \mathbb{R}$:

\begin{equation}
S\left(\left[\begin{array}{c} v \\ \omega\end{array}\right]\right) \left[\begin{array}{c} p \\ q\end{array}\right]= \left[\begin{array}{c} vq + \omega \times p \\ q\end{array}\right]. 
\end{equation}

If $H[k]$ is the pose of object $\mathcal{B}$ at step $k$, given parameters $(\Delta t, \beta, \kappa, t_{max})$, then the Nesterov update can be written as:

\begin{eqnarray}
\label{eq:nesterov_update}
    && \tilde{H} = \exp\big(\beta \Delta t S(\xi[k])\big)H[k] \nonumber \\
    &&\xi[k+1] = \beta \xi[k] + \kappa L_{\mathcal{B}}[f](\tilde{H})^{\top} \nonumber \\
    && H[k+1] =  \exp\big(\Delta t S(\xi[k+1])\big)H[k]
\end{eqnarray}

\noindent in which $H[0]$ starts with the initial pose and $\xi[0]$ as the zero vector. This updates run up until there is a separation (validated through Euclidean distance computation) or $ k \Delta t > t_{max}$.

The parameters selected for each algorithm are shown in Table \ref{tab:escape_parameters}, together with the selected smoothng factor $h$. The parameters were selected empirically through trial and error, with the aim of maximizing the success rate of each method in achieving geometrically certified separation. Once selected, these parameters were kept
unchanged across all configurations. The results can be seen in Table \ref{tab:escape_comparison}, in which 40,000 pairs of objects with random initial poses were sampled. The objects were taken from the TetWild dataset \cite{hu2018tetwild} and had their surfaces sampled to obtain discrete points that were used in each algorithm. Both algorithms were evaluated on every object pair and initial configuration. The results indicate that $h$-NESDF can perform separations more often than the algorithm in ``\cite{beker2025smooth}''.

\begin{table}[!t]
    \centering
    \caption{Parameters used in the separation experiments.}
    \label{tab:escape_parameters}
    \footnotesize
    \setlength{\tabcolsep}{3pt}
    \renewcommand{\arraystretch}{1.1}
    \begin{tabular}{lcc}
        \hline
        \textbf{Parameter}
        & \textbf{$h$-NESDF}
        & \textbf{Beker}\\
        \hline
        Smoothing $h$ & $0.1$ & 0.06\\
        Gain $\kappa$ & $2.0$ & $0.5$\\
        Momentum $\beta$ & $0.3$ & $0.3$\\
        Step $\Delta t$ (s) & $0.2$ & $0.1$\\
        Time limit (s) & $20$ & $20$\\
        \hline
    \end{tabular}
\end{table}

\begin{table}[!t]
    \centering
    \caption{Geometrically certified separation results over $40,000$ shared initial configurations.}
    \label{tab:escape_comparison}
    \footnotesize
    \renewcommand{\arraystretch}{1.15}
    \begin{tabular}{lcc}
        \hline
        \textbf{Metric} & \textbf{$h$-NESDF} & \textbf{Beker}\\
        \hline
        Successful  separations & 36,410 & 28,124\\
        Unsuccessful separations  & 3,590  & 11,876\\
        Rate (\%)   & 91.03  & 70.31\\
        \hline
    \end{tabular}
\end{table}

\subsection{Computational Efficiency}

We evaluate the computational efficiency of the $h$-NESDF algorithm, approximated as discussed in Subsection \ref{subs:implpractice}, when the $h$-ESDF for $\mathcal{A}$ is computed using a point cloud, as described in Subsection \ref{subs:compute_D}. In this case, assuming $\mathcal{A}$ and $\mathcal{B}$ have $K_{\mathcal{A}}$ and $K_{\mathcal{B}}$ points, respectively, the computational complexity of the algorithm is of order $\mathcal{O}(K_{\mathcal{A}}K_{\mathcal{B}})$, since the $h$-NESDF makes $K_{\mathcal{B}}$ calls to the $h$-ESDF of $\mathcal{A}$, which in turn has a complexity of $\mathcal{O}(K_{\mathcal{A}})$. In comparison, as mentioned in  Subsection \ref{subs:qualcomp}, \cite{beker2025smooth} uses a similar implementation whose computational complexity, when applied to point clouds, is also $\mathcal{O}(K_{\mathcal{A}}K_{\mathcal{B}})$.

The $h$-NESDF code was implemented in PyTorch using FP32 precision on an NVIDIA T4 GPU. The shapes were chosen randomly: because the algorithm is explicit rather than iterative, the computational time depends primarily on the number of points in each shape, apart from minor random fluctuations. 

\begin{figure}[!t]
    \centering
    \includegraphics[width=0.5\textwidth]{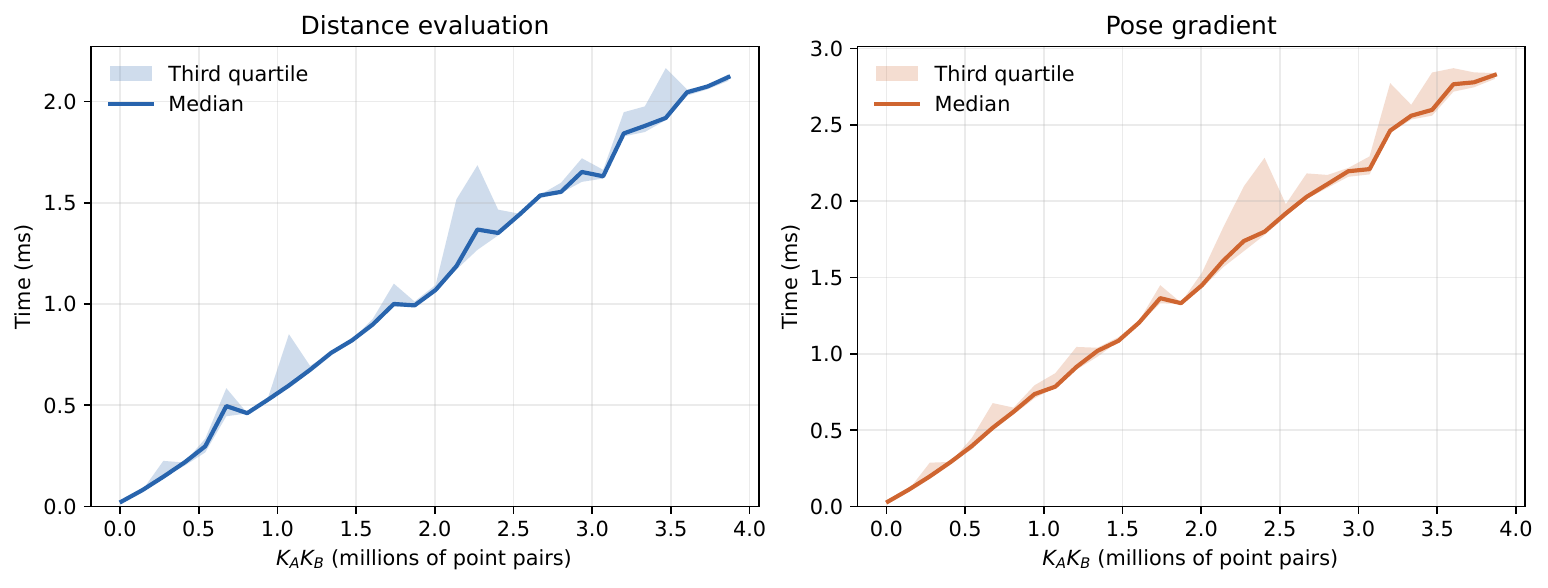}
    \caption{Computational time for the $h$-NESDF and the additional time required to compute its pose gradient.}
    \label{fig:nesdf_timings}
\end{figure}

We tested $6,000$ pairs with $K_{\mathcal{A}} = K_{\mathcal{B}}$ ranging from $100$ to $2,000$. For each pair, we computed $N_{\mathcal{A},\mathcal{B}}^h$ as well as its gradient with respect to the poses of both objects. More precisely, let $N^h_{\mathcal{A},\mathcal{B}}(Q_{\mathcal{A}},s_{\mathcal{A}},Q_{\mathcal{B}},s_{\mathcal{B}})$ denote the $h$-NESDF as a function of the object poses $(Q_{\mathcal{A}},s_{\mathcal{A}})$ and $(Q_{\mathcal{B}},s_{\mathcal{B}})$. We compute the \textsl{SE}(3) gradient-like quantity $\textsl{L}[N^h_{\mathcal{A},\mathcal{B}}](Q_{\mathcal{A}},s_{\mathcal{A}},Q_{\mathcal{B}},s_{\mathcal{B}})$ as in \eqref{eq:escape_operator_full}. This gradient was computed using PyTorch's automatic differentiation. We then aggregated the execution times based on the product $K_{\mathcal{A}}K_{\mathcal{B}}$. Figure \ref{fig:nesdf_timings} presents the results in milliseconds per distance computation, showing the median (second quartile) alongside the third quartile. 

The GPU implementation also supports batching to compute distances for multiple pairs in parallel. To avoid out-of-memory errors, the batch size $k_{p}$ was determined empirically as $k_{p} = \textsl{round}(4\times 10^{7}/(K_{\mathcal{A}}K_{\mathcal{B}}))$.

\subsection{Motion Planning Application}

\begin{figure}[!t]
    \centering
    \includegraphics[width=0.3\textwidth]{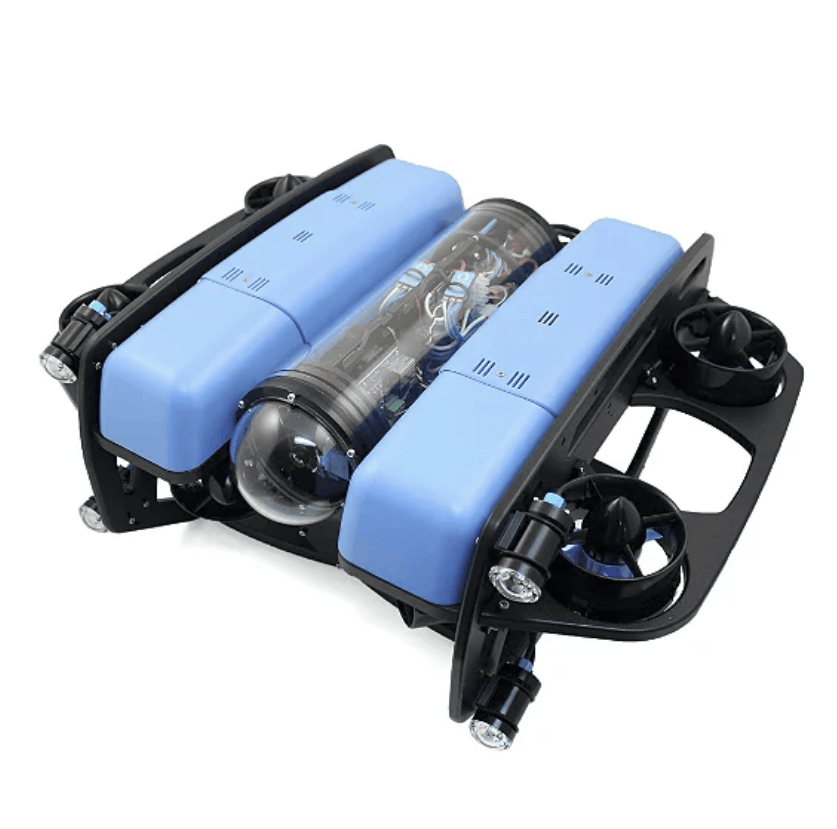}
    \caption{The BlueROV2 in its Heavy configuration. Its eight-thruster
    arrangement provides actuation in all six degrees of
    freedom~\cite{bluerobotics_bluerov2}.}
    \label{fig:bluerov2}
\end{figure}

\begin{figure*}[!t]
    \centering
    \includegraphics[width=0.3\textwidth]{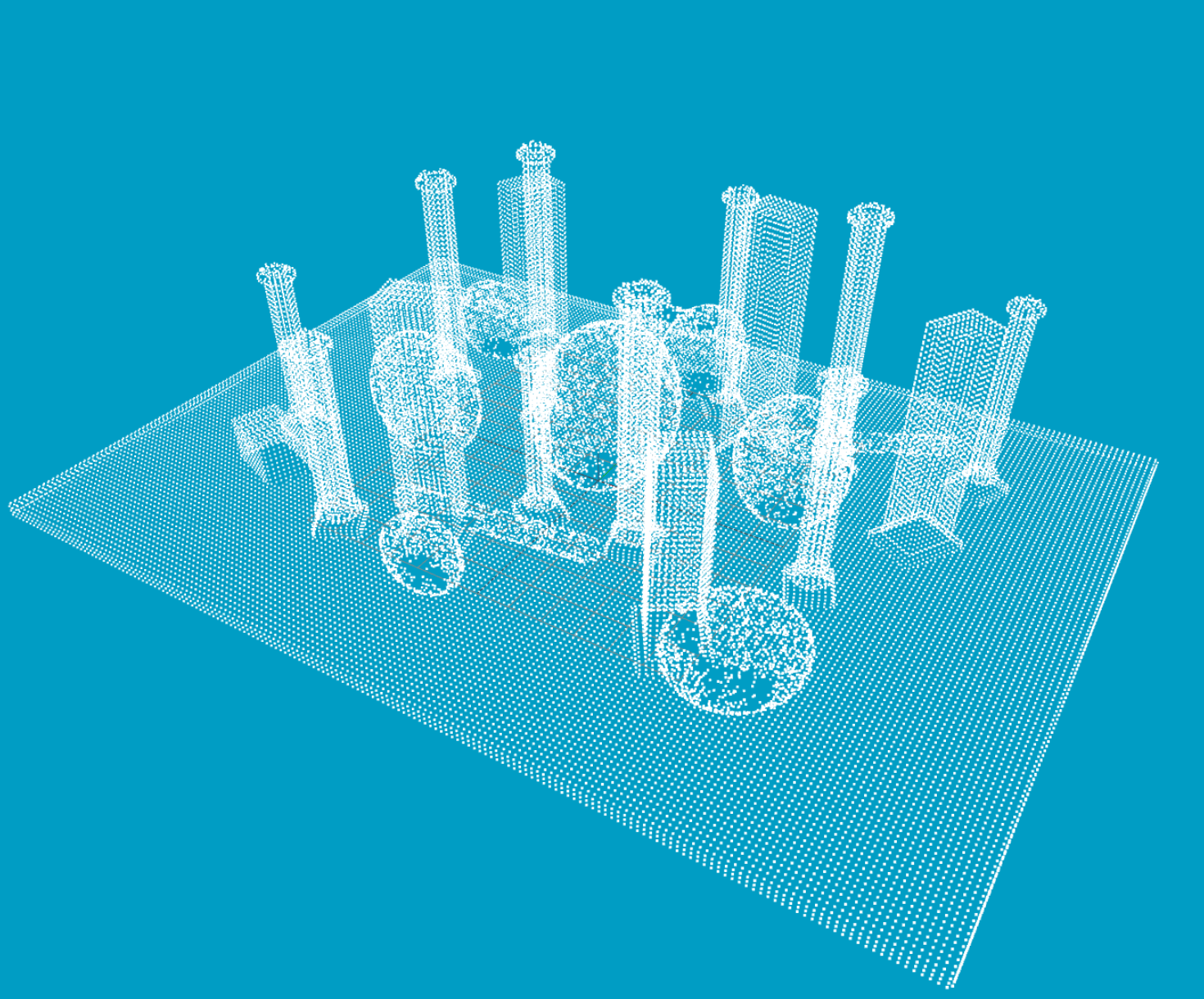}
    \includegraphics[width=0.3\textwidth]{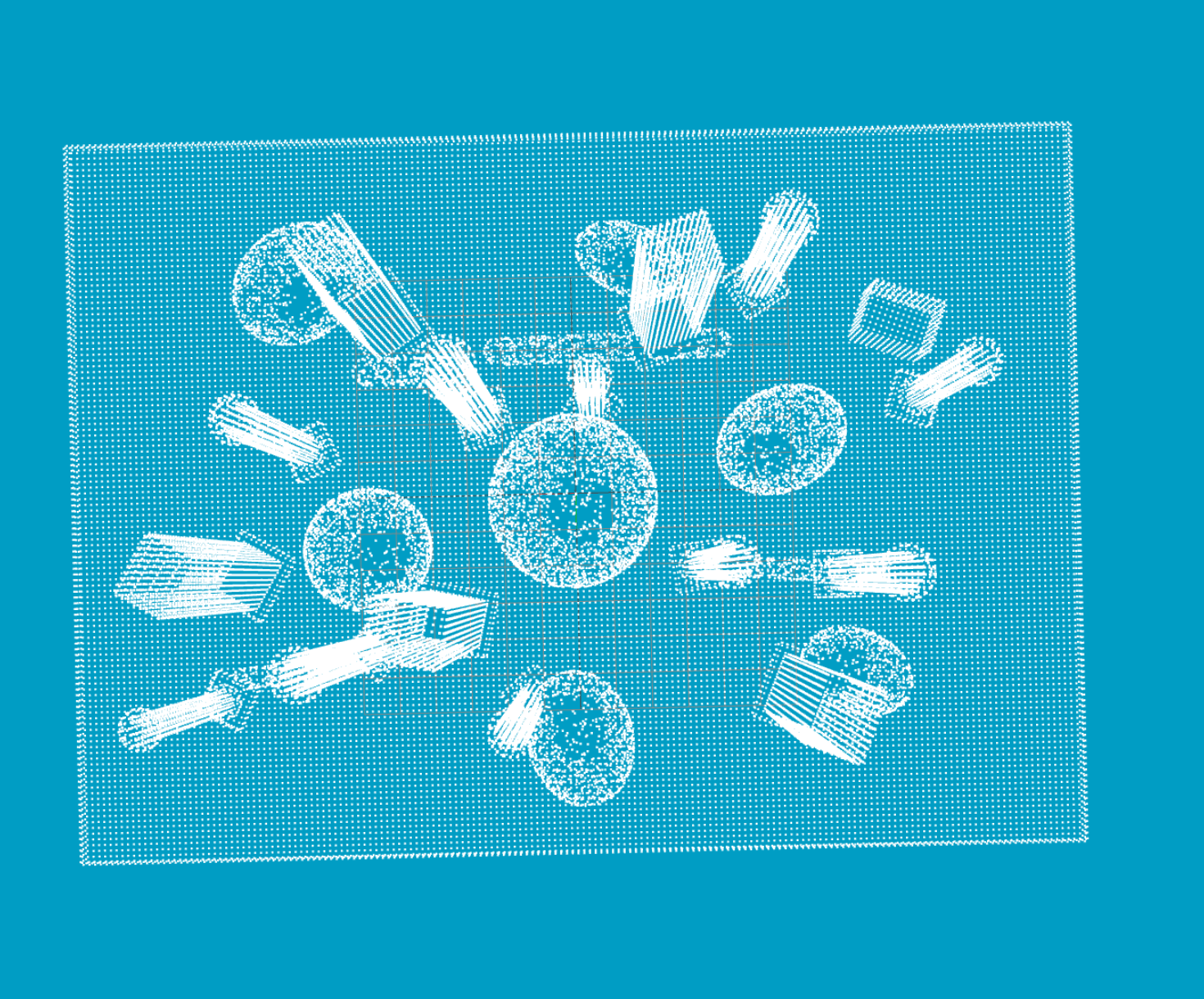}
    \includegraphics[width=0.3\textwidth]{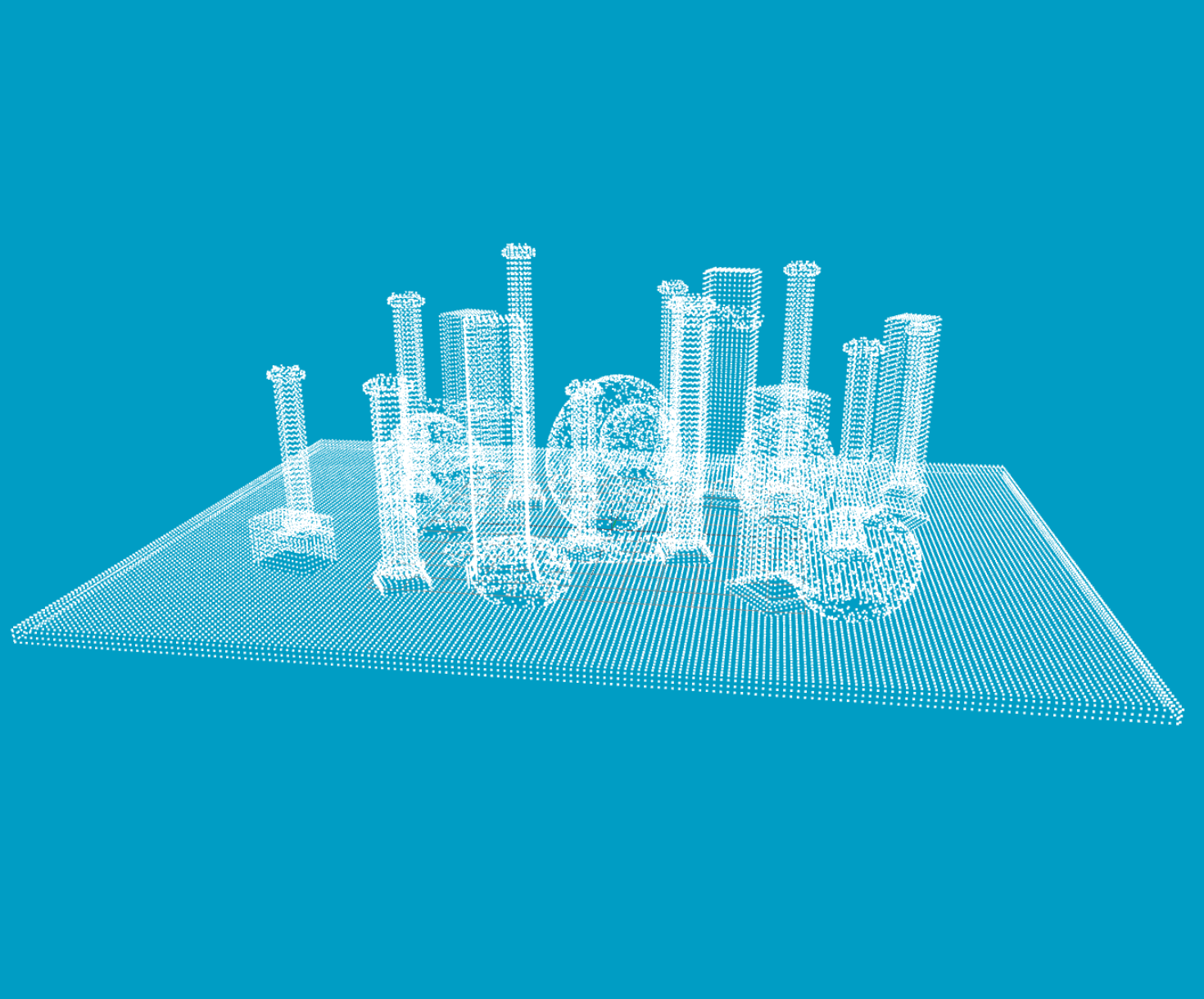}
    \caption{Synthetic underwater environment containing rocks, boxes,
    and vertical pipes, represented by an oriented point cloud with
    $54{,}570$ points.}
    \label{fig:underwater_environment}
\end{figure*}

To evaluate the applicability of the $h$-NESDF to motion planning, we
constructed a synthetic underwater environment represented by an oriented
point cloud, as shown in Figure~\ref{fig:underwater_environment}. We
consider a fully actuated underwater vehicle with independent control of
its three translational and three rotational degrees of freedom, such as
the BlueROV2 in its Heavy configuration
(Figure~\ref{fig:bluerov2})~\cite{bluerobotics_bluerov2}. For geometric
planning, the vehicle is modeled as a rigid body with a box-shaped
approximation of its geometry.

Given initial and goal poses
$H_{\mathrm{start}},H_{\mathrm{goal}}\in\mathrm{SE}(3)$, the motion-planning
problem is to find a continuous path $H:[0,1]\to\mathrm{SE}(3)$ connecting
these poses while avoiding collisions with the environment. We use a
CHOMP-inspired trajectory optimization scheme~\cite{Ratliff2009}. The
path is discretized into $T$ segments, with $T+1$ poses
$H[0],\ldots,H[T]$. The endpoints are fixed at
$H[0]=H_{\mathrm{start}}$ and $H[T]=H_{\mathrm{goal}}$, and the intermediate
poses $H[1],\ldots,H[T-1]\in\mathrm{SE}(3)$ are optimized by maximizing
$f(H[1],\ldots,H[T-1])$, defined by
\begin{equation}
\begin{aligned}
    f={}&-\sum_{k=0}^{T-1}\Delta\bigl(H[k],H[k+1]\bigr)\\
       &-C_{\mathrm{weight}}h\ln\!\left(
           \sum_{k=0}^{T}e^{-N_h(H[k])/h}
       \right).
\end{aligned}
\label{eq:motion_objective}
\end{equation}
Here, $N_h(H)$ denotes the $h$-NESDF between the box $\mathcal A$
representing the vehicle at pose $H$ and the environment $\mathcal B$,
represented by its oriented point cloud. The second term rewards a
log-sum-exp approximation of the minimum clearance along the discretized
path, with smoothing parameter $h>0$ and weight
$C_{\mathrm{weight}}>0$.

The first term penalizes excessive separation between consecutive poses.
For poses $H=(Q,s)$ and $H'=(Q',s')$, define the weighted pose distance
\begin{equation}
    d_{\mathrm{pose}}(H,H')^2
    =\|s'-s\|_2^2
     +C_{\mathrm{factor}}\|Q'-Q\|_F^2,
\label{eq:motion_pose_distance}
\end{equation}
where $s,s'\in\mathbb R^3$, $Q,Q'\in\mathrm{SO}(3)$, and
$C_{\mathrm{factor}}>0$ balances the translational and rotational
contributions. We use the barrier penalty
$\Delta:\mathrm{SE}(3)\times\mathrm{SE}(3)\to\mathbb R\cup\{+\infty\}$,
defined, with $r=d_{\mathrm{pose}}(H,H')$, by
\begin{equation}
    \Delta(H,H')=
    \begin{cases}
    -\ln\!\left(1-
        \dfrac{r^2}{C_{\mathrm{threshold}}^2}\right),
        & r<C_{\mathrm{threshold}},\\[6pt]
    +\infty, & \text{otherwise}.
    \end{cases}
\label{eq:motion_barrier}
\end{equation}
Thus, consecutive poses must satisfy the strict spacing bound
$d_{\mathrm{pose}}(H[k],H[k+1])<C_{\mathrm{threshold}}$.

The smooth box field $D_{\mathcal A}^h(b)$ is constructed from the exact
signed Euclidean distance to a box.  Let $Q,s$ represent the current pose of the box $\mathcal{A}$. Let
$\ell=[\ell_x \ \ell_y \ \ell_z]^\top/2$ be its half-length vector and define
$u=Q^\top(b-s)$ and $v=|u|-\ell$. The exact distance is
\begin{equation}
    D_{\mathcal A}(b)
    =\|\max(v,0)\|_2+\min\!\left\{\max_i v_i,0\right\},
\label{eq:motion_box_esdf}
\end{equation}
where the absolute value and $\max(v,0)$ are applied componentwise.
We replace maxima and minima by their log-sum-exp approximations with
smoothing parameter $h$, expressing $|u_i|$ as $\max\{u_i,-u_i\}$. We also
regularize the outer norm as
$\sqrt{\|\cdot\|_2^2+\epsilon_h^2}$, with
$\epsilon_h=h^2/\|\ell\|_2$. These substitutions yield a $h$-ESDF $E_{\mathcal{A}}^h(b)$ for the query point $b$ and box pose for every
$h>0$.

Optimization proceeds by gradient ascent on $f$, using the Lie-group
differential $L[f]$ introduced in
Subsection~\ref{subsec:escape_comparison}, evaluated with respect to each
free pose $H[k]$. The poses are updated using the Nesterov scheme in
\eqref{eq:nesterov_update}, while the endpoints remain fixed and updates
are restricted to the feasible domain of the spacing barrier.

We performed $200$ motion-planning trials with randomly sampled pairs of
initial and goal poses. Both endpoints were required to be collision-free
and sufficiently far apart. In addition, the interpolated initial path
was required to intersect the environment, ensuring that each trial
required a modification of the initial path to achieve collision
avoidance, and thus was not trivial. The initial path was generated by linear interpolation of the
translation and shortest-arc quaternion spherical linear interpolation
(SLERP) of the rotation~\cite{shoemake1985animating}.
Table~\ref{tab:motion_parameters} lists the parameters used in these
experiments.

\begin{table}[!t]
    \centering
    \caption{Parameters used in the motion-planning experiments.}
    \label{tab:motion_parameters}
    \footnotesize
    \setlength{\tabcolsep}{3pt}
    \renewcommand{\arraystretch}{1.1}
    \begin{tabular}{lc}
        \hline
        \textbf{Parameter} & \textbf{Value} \\
        \hline
        $h$ & $0.1$ \\
        $T$ & $30$ \\
        $C_{\mathrm{weight}}$ & $25$ \\
        $C_{\mathrm{factor}}$ & $1$ \\
        $C_{\mathrm{threshold}}$ & $0.6$ \\
        Momentum $\beta$ & $0.95$ \\
        Step size $\Delta t$ & $0.2$ \\
        $\kappa$ & $0.5$ \\
        \hline
    \end{tabular}
\end{table}

\begin{figure}[!t]
    \centering

    \includegraphics[width=0.45\linewidth]{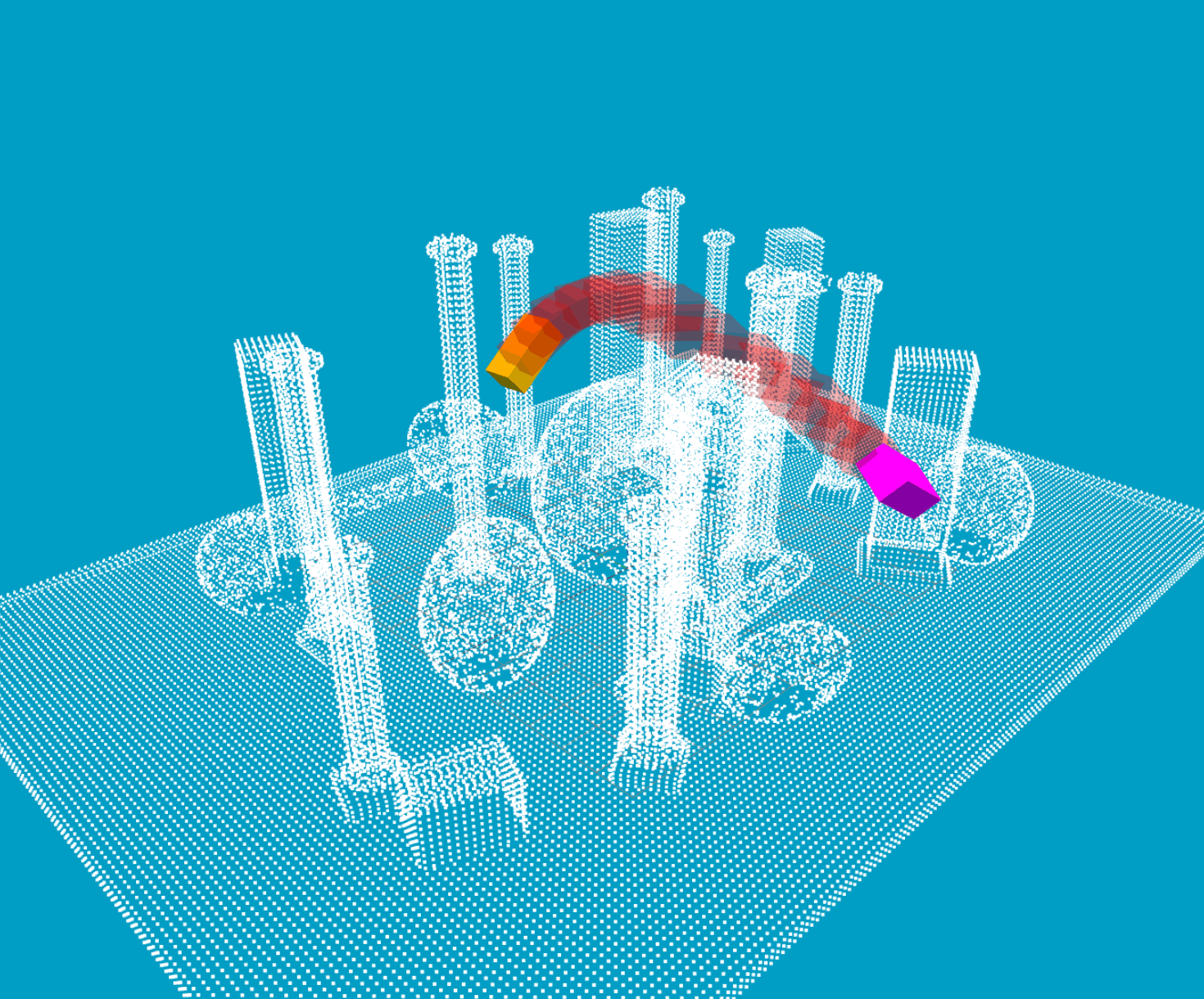}
    \hfill
    \includegraphics[width=0.45\linewidth]{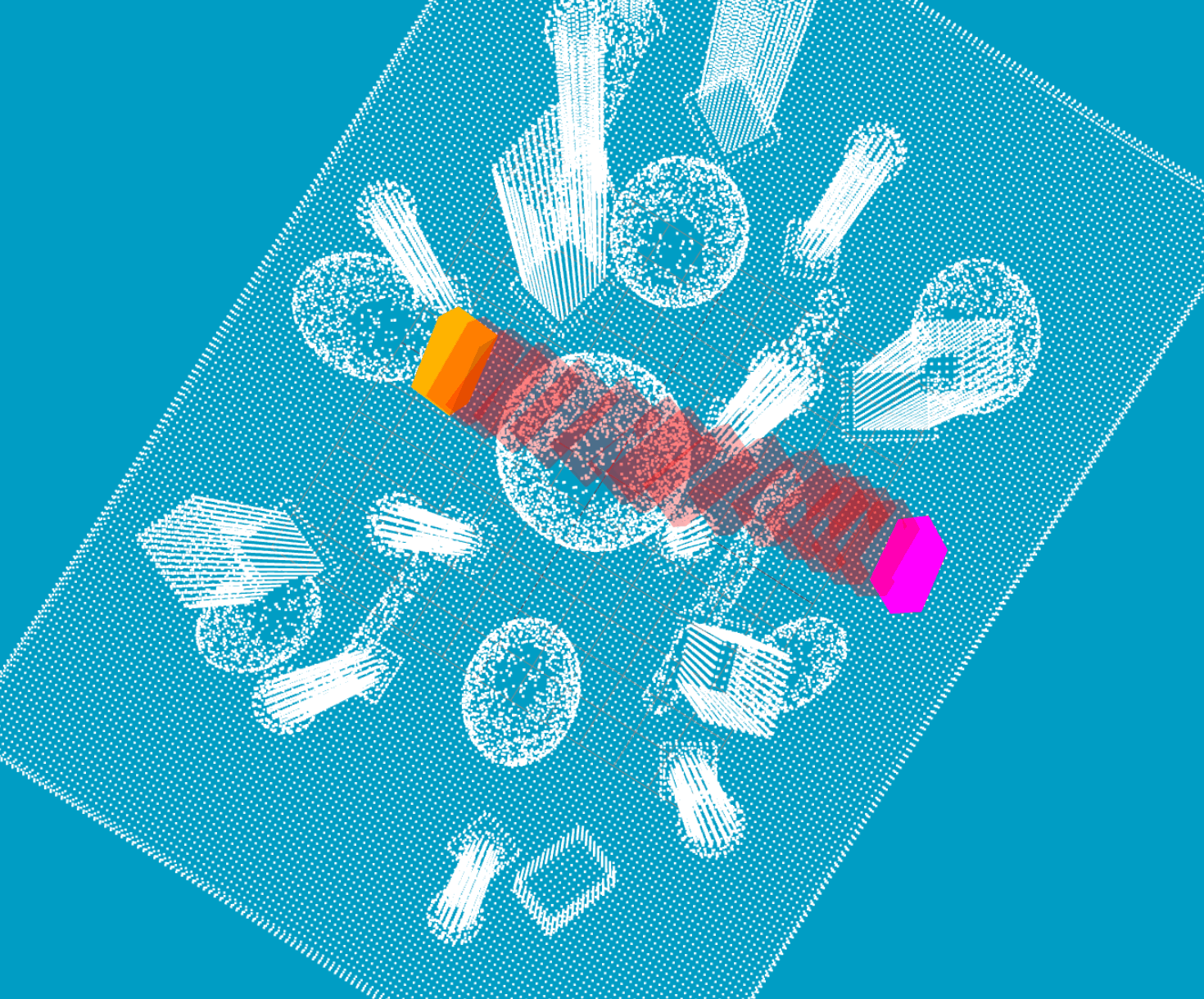}

    \par\medskip

    \includegraphics[width=0.45\linewidth]{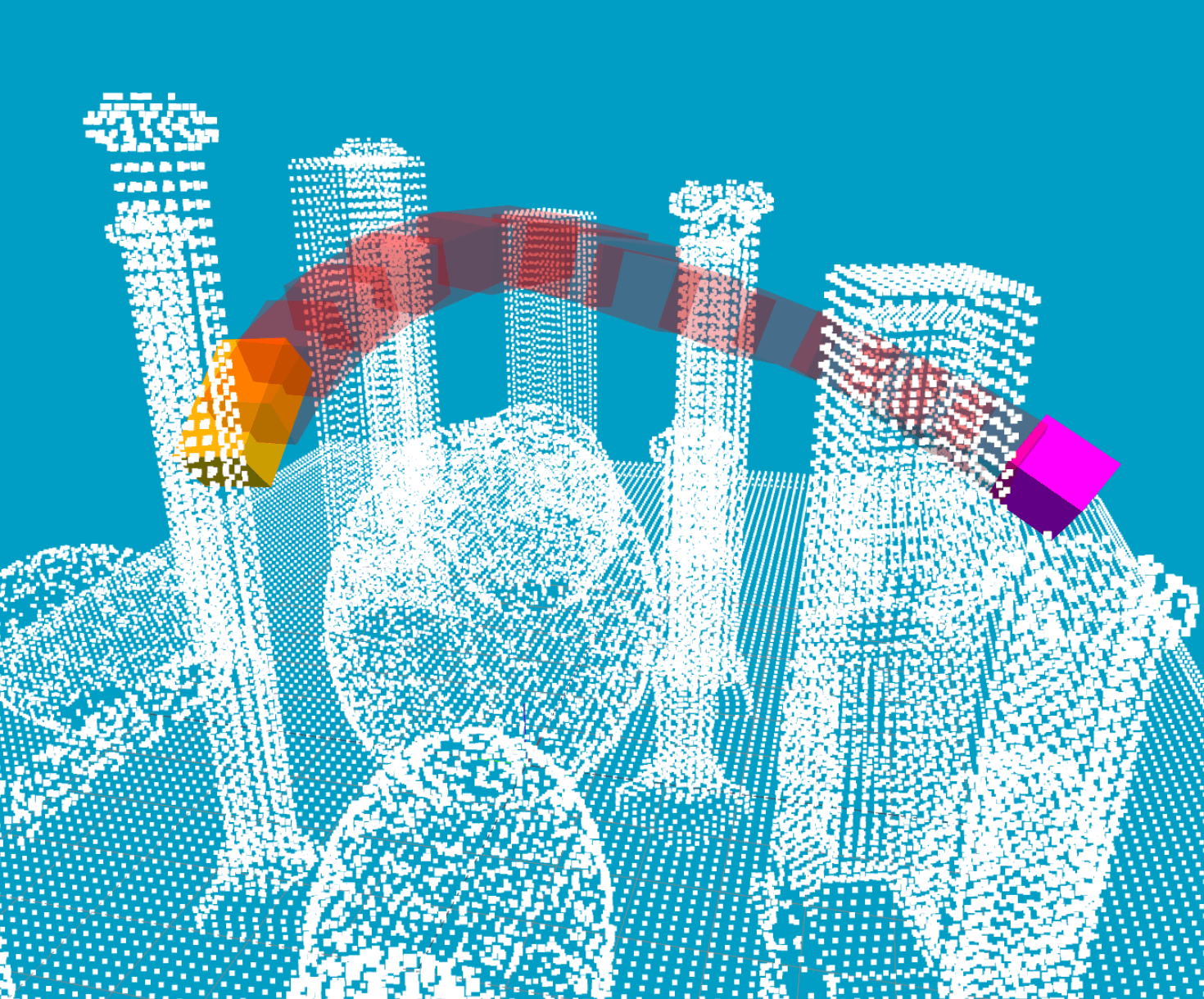}
    \hfill
    \includegraphics[width=0.45\linewidth]{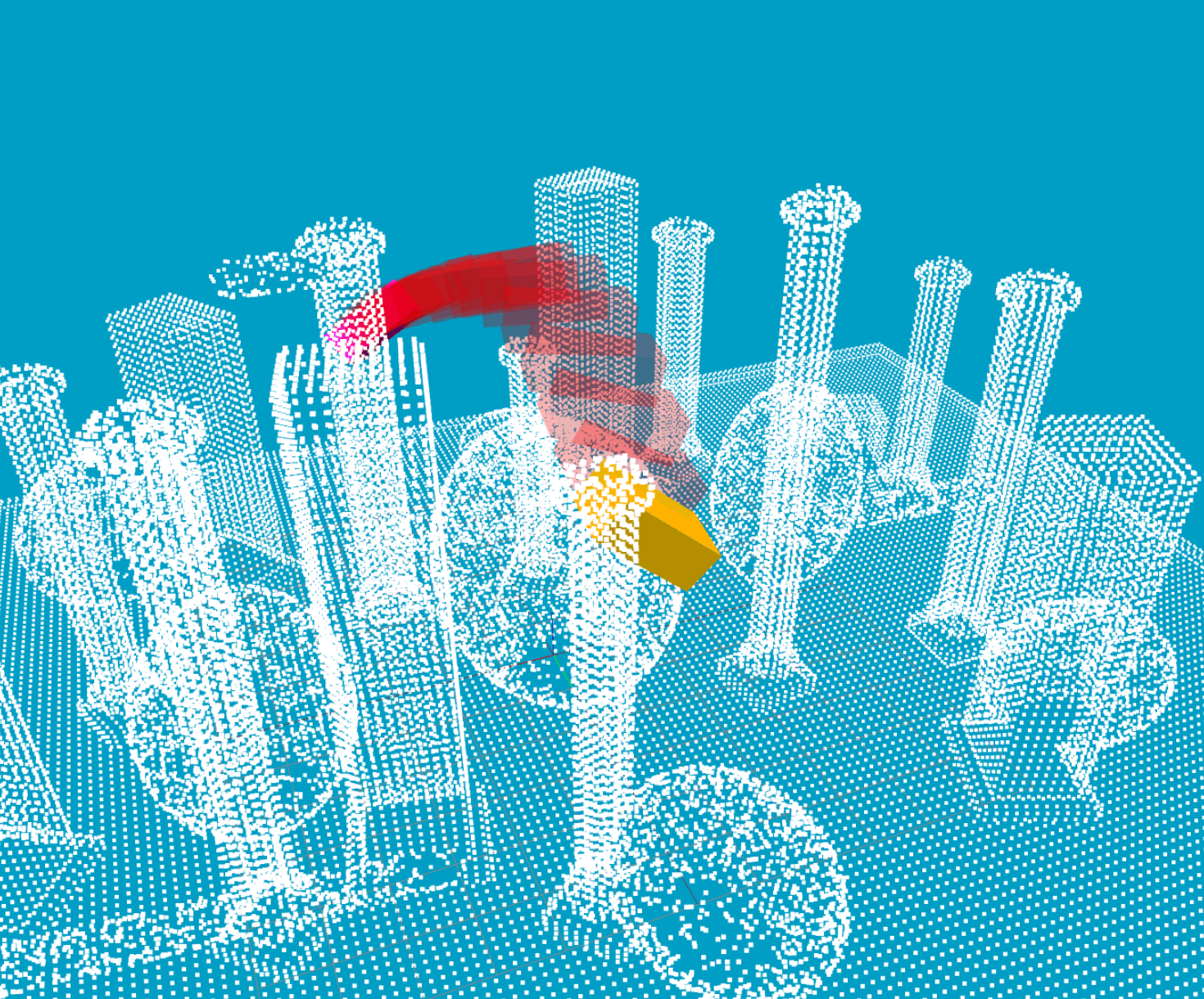}

    \caption{Example of a motion planning performed by the CHOMP-like algorithm.}
    \label{fig:underwater_panels}
\end{figure}

The algorithm was implemented in a NVIDIA T4 GPU using batch calculation for the distance between the object at the $T+1$ poses $H[k]$ and the point cloud environment. Figure \ref{fig:underwater_panels} shows an example of result obtaind from this algorithm, in four different views. The yellow box is the starting pose, the magenta box the final pose, and the red trail the optimized trajectory.

In the final result, $34\%$ of the $200$ tests returned success, with an average computational time of $0.7s$.

\section{Conclusion}

In this paper, we introduced the $h$-NESDF, a signed and
infinitely differentiable distance function between two sets
satisfying the stated geometric assumptions. We established
theoretical properties, provided guidelines for its approximate
implementation, and evaluated its computational efficiency
and applicability in robotic, especially motion planning, through numerical experiments. Future work will focus on the following
directions:

\begin{enumerate}
    \item The $h$-NESDF does not guarantee positivity if and
    only if the two objects are disjoint, although this
    property is desirable in many applications.
    Reference~\cite{bartelt2026h} introduced an almost
    everywhere smooth approximation of the minimum operator,
    denoted by $\min[h]$, that preserves the sign of the
    exact minimum, that is,  $\operatorname{sign}\!\left(\min[h]_i x_i\right)  =  \operatorname{sign}\!\left(\min_i x_i\right)$.
    Such an operator may provide a useful ingredient for
    developing a formulation with exact collision-detection
    guarantees or, at least, like in \cite{bartelt2026h} , one that can be controlled independently of the smoothing parameter $h$ at the expense of computational time. Its potential integration into the
    $h$-NESDF framework will be investigated in future work.

    \item Direct evaluation of the point-cloud approximation
    has computational complexity
    $\mathcal{O}(K_{\mathcal{A}}K_{\mathcal{B}})$.
    This cost can become prohibitive for dense point clouds
    and applications requiring low-latency distance queries, even when GPUs and batching are used.
    However, the exponential weighting suggests opportunities
    for acceleration. Indeed, if $b_k$ are the points sampled from object  $\mathcal{B}$  and $\tilde{d} = \min_k E_{\mathcal{A}}^h(b_k)$ , if $(E_{\mathcal{A}}^h(b_k)-\tilde{d})/(h^2)$ is sufficiently large , this point $b_k$  influence very little in the final calculation.  A similar
    observation applies to the evaluation of
    $E_{\mathcal{A}}^h(b_k)$ itself. These observations
    motivate investigating spatial data structures, such as
    k-d trees, to identify and omit negligible contributions
    without evaluating every point pair. Developing an
    efficient GPU implementation requires careful management
    of memory access and thread divergence, together with
    control of the accumulated approximation error and
    preservation of smoothness.
\end{enumerate}
```

\section{Appendix}

\begin{lemma} \label{lemma:kkt} Let $(\mathcal{A},\mathcal{B})$ be a nondegenerate pair. Then there is a scalar $\lambda$ such that
\begin{equation}
  \nabla E_{\mathcal{A}}(b^*) = \lambda n_{\mathcal{B}}(b^*).  
\end{equation}
\noindent Furthermore, $|\lambda| = D_{\mathcal{A}}(b^*)-d_{\mathcal{A}}$.
    
\end{lemma}

\begin{proof}
    Consider the Karush--Kuhn--Tucker (KKT) conditions for the optimization problem defining $b^*$:
\begin{align}
\label{eq:optprob}
    \min_{b} E_{\mathcal{A}}(b) \ \mbox{such that} \ D_{\mathcal{B}}(b)=0.
\end{align}
Since $n_{\mathcal{B}}(b) = \nabla D_{\mathcal{B}}(b)$ when $b \in \partial \mathcal{B}$, the KKT stationarity condition tells us that there is a $\lambda \in \mathbb{R}$ such that $\nabla E_{\mathcal{A}}(b^*) = \lambda n_{\mathcal{B}}(b^*)$.

Now, note that $\nabla E_{\mathcal{A}}(b) = (D_{\mathcal{A}}(b)-d_{\mathcal{A}})\nabla D_{\mathcal{A}}(b)$. Since $D_{\mathcal{A}}(b)$ is a signed distance, whenever $\nabla D_{\mathcal{A}}(b)$ exists, $\|\nabla D_{\mathcal{A}}(b)\|=1$ (the \emph{Eikonal equation} \cite{OsherFedkiw2003}). Thus, if $\nabla D_{\mathcal{A}}(b^*)$ exists,

\begin{equation}
    \|\nabla E_{\mathcal{A}}(b^*)\| = D_{\mathcal{A}}(b^*)-d_{\mathcal{A}}
\end{equation}

\noindent where we used the fact that $D_{\mathcal{A}}(b)-d_{\mathcal{A}} \geq 0$ for all $b$. Finally, taking norms on both sides of $\nabla E_{\mathcal{A}}(b^*) = \lambda n_{\mathcal{B}}(b^*)$ and using $\|n_{\mathcal{B}}(b^*)\|=1$ gives the desired result.
$\blacksquare$
\end{proof}

\begin{lemma} \label{lemma:twocases}Let $(\mathcal{A},\mathcal{B})$ be a nondegenerate pair. Then:
    \begin{enumerate}
        \item \textbf{Case 1}: If $\textsl{int}(\mathcal{A}) \cap \textsl{int}(\mathcal{B}) = \emptyset$, then $N_{\mathcal{A},\mathcal{B}} = D_{\mathcal{A}}(b^*)$. 
        \item \textbf{Case 2}: If $\mathcal{A} \subseteq \mathcal{B}$, then $N_{\mathcal{A},\mathcal{B}} = 2d_{\mathcal{A}}-D_{\mathcal{A}}(b^*)$.
    \end{enumerate}
\end{lemma}

\begin{proof}

In \textbf{Case 1}, we claim that we can replace the constraint $D_{\mathcal{B}}(b)=0$ by $D_{\mathcal{B}}(b)\leq0$ without changing the minimizer. Suppose, for a contradiction, that a minimizer $b^*$ of the relaxed problem satisfies $D_{\mathcal{B}}(b^*)<0$, that is, $b^* \in \textsl{int}(\mathcal{B})$. Since the constraint is inactive, $b^*$ must also be a local minimum of the objective function $E_{\mathcal{A}}(b)$. But $\textsl{int}(\mathcal{A}) \cap \textsl{int}(\mathcal{B}) = \emptyset$, so $b^* \notin \textsl{int}(\mathcal{A})$. This is impossible: for a regular set, neither the signed distance nor $E_{\mathcal{A}}$ has a local minimum outside its interior. Hence, the relaxed minimizer lies on $\partial\mathcal{B}$ and coincides with the original $b^*$ by uniqueness. Dual feasibility in the KKT conditions then implies that the Lagrange multiplier $\lambda$ is nonpositive. Together with Lemma \ref{lemma:kkt}, this gives $\nabla E_{\mathcal{A}}(b^*) = -(D_{\mathcal{A}}(b^*)-d_{\mathcal{A}}) n_{\mathcal{B}}(b^*)$. Substituting this into the definition of $N_{\mathcal{A},\mathcal{B}}$ gives the desired result.

\textbf{Case 2} is similar, but now we replace the constraint $D_{\mathcal{B}}(b)=0$ by $D_{\mathcal{B}}(b)\geq0$. Suppose, for a contradiction, that a minimizer $b^*$ of the relaxed problem satisfies $D_{\mathcal{B}}(b^*)>0$, that is, $b^* \notin \mathcal{B}$. Since the constraint is inactive, $b^*$ must be a local minimum of $E_{\mathcal{A}}(b)$. This is impossible: $\mathcal{A}\subseteq\mathcal{B}$ implies that $b^*\notin\mathcal{A}$, whereas $E_{\mathcal{A}}$ has no local minimum outside $\mathcal{A}$. Hence, the relaxed minimizer lies on $\partial\mathcal{B}$ and coincides with the original $b^*$ by uniqueness. Dual feasibility in the KKT conditions then implies that $\lambda$ is nonnegative. Together with Lemma \ref{lemma:kkt}, this gives $\nabla E_{\mathcal{A}}(b^*) = (D_{\mathcal{A}}(b^*)-d_{\mathcal{A}}) n_{\mathcal{B}}(b^*)$. Substituting this into the definition of $N_{\mathcal{A},\mathcal{B}}$ gives the desired result.
$\blacksquare$
\end{proof}

\textbf{Proof of Proposition 1}: For \textbf{Cases 1 and 2}, apply Case 1 of Lemma \ref{lemma:twocases} to obtain $N_{\mathcal{A},\mathcal{B}} = D_{\mathcal{A}}(b^*)$. If $\partial \mathcal{A} \cap \partial \mathcal{B} = \emptyset$, then $D_{\mathcal{A}}(b^*)>0$; if $\partial \mathcal{A} \cap \partial \mathcal{B} \neq \emptyset$, then $D_{\mathcal{A}}(b^*)=0$.

\textbf{Case 3} requires two subcases. In addition to $\textsl{int}(\mathcal{A}) \cap \textsl{int}(\mathcal{B}) \neq \emptyset$, either \textbf{Case 3.1} $\textsl{int}(\mathcal{A}) \cap \partial \mathcal{B} \neq \emptyset$ or \textbf{Case 3.2} $\textsl{int}(\mathcal{A}) \cap \partial \mathcal{B} = \emptyset$ holds.

In \textbf{Case 3.1}, note that $D_{\mathcal{A}}(b^*)<0$. But

\begin{eqnarray}
    N_{\mathcal{A},\mathcal{B}} &\leq& d_{\mathcal{A}} + |n_{\mathcal{B}}(b^*)^{\top} \nabla E_{\mathcal{A}}(b^*)| \nonumber \\
    &\leq & d_{\mathcal{A}} +\|n_{\mathcal{B}}(b^*)\|\| \nabla E_{\mathcal{A}}(b^*)\| \nonumber \\
    &=& d_{\mathcal{A}}+\big(D_{\mathcal{A}}(b^*)-d_{\mathcal{A}} \big) = D_{\mathcal{A}}(b^*)
\end{eqnarray}

\noindent and so $N_{\mathcal{A},\mathcal{B}} < 0$. It remains to establish \textbf{Case 3.2}, where $\textsl{int}(\mathcal{A}) \cap \partial \mathcal{B} = \emptyset$. Since $\textsl{int}(\mathcal{A})$ is path connected and $\mathcal{A}=\overline{\textsl{int}(\mathcal{A})}$, the conditions $\textsl{int}(\mathcal{A}) \cap \partial \mathcal{B} = \emptyset$ and $\textsl{int}(\mathcal{A}) \cap \textsl{int}(\mathcal{B}) \neq \emptyset$ imply $\mathcal{A} \subseteq \mathcal{B}$. We can therefore apply Case 2 of Lemma \ref{lemma:twocases}: $N_{\mathcal{A},\mathcal{B}} = 2d_{\mathcal{A}} - D_{\mathcal{A}}(b^*)$. This is negative because $d_{\mathcal{A}} < 0$ and $D_{\mathcal{A}}(b) \geq d_{\mathcal{A}} > 2d_{\mathcal{A}}$ for all $b$.

$\blacksquare$.

\textbf{Proof of Proposition 2}: In this case, $d_{\mathcal{A}} = 0$, $E_{\mathcal{A}}(b) = 0.5\|b-a\|^2$, and thus
    \begin{equation}
    b^* = \arg \min_{b \in \partial \mathcal{B}} \|b-a\| \ , \ N_{\mathcal{A},\mathcal{B}} = n_{\mathcal{B}}(b^*)^{\top}(a-b^*)
    \end{equation}
    For a smooth boundary and a unique closest point, this is the ESDF between the point $a$ and the set $\mathcal{B}$ \cite{BaerentzenAanaes2005}. $\blacksquare$

\begin{lemma} \label{lemma:lemma3}
    On any compact set $\mathcal{U} \subseteq \partial \mathcal{B}$, define the measure:
    \begin{equation}
    \label{eq:muh}
 \mu_h(\mathcal{U}) \triangleq  \int_{\mathcal{U}} \rho_{\mathcal{A},h}(\beta)dS
\end{equation}
\noindent in which $\rho_{\mathcal{A},h}$ is the density defined in \eqref{eq:rho}. If $(\mathcal{A},\mathcal{B})$ is a non-degenerate pair, for any open set $\mathcal{N} \subset \partial \mathcal{B}$ that contains $b^*$ (as defined in \eqref{eq:bstar}), there exist positive constants $\Delta_{\mathcal{N}}$ and $C_{\mathcal{N}}$ such that, for $h$ sufficiently small:
\begin{equation}
    \mu_h(\partial \mathcal{B}-\mathcal{N}) \leq C_{\mathcal{N}}e^{-\Delta_{\mathcal{N}}/(2h^2)}.
\end{equation}
\end{lemma}
\begin{proof} 
Define, for any open set $\mathcal{N}$:

\begin{equation}
\label{eq:DeltaN}
    \Delta_{\mathcal{N}} \triangleq \min_{b \in \partial \mathcal{B}-\mathcal{N}} E_{\mathcal{A}}(b) - E_{\mathcal{A}}(b^*).
\end{equation}

Note that if $b^* \in \mathcal{N}$  then necessarily $\Delta_{\mathcal{N}} > 0$. This comes from the fact that $b^*$ minimizes $E_{\mathcal{A}}(b)$ over $\partial \mathcal{B}$ and it is the \emph{unique} minimizer (this comes from the fact that the pair ($\mathcal{A},\mathcal{B}$) is nondegenerate). 

Define $\sigma(\mathcal{U}) = \int_{\mathcal{U}} dS$. Now, still considering an open $\mathcal{N} \subseteq \partial \mathcal{B}$ such that $b^* \in \mathcal{N}$,   pick the set $\mathcal{M}$ with maximum $\sigma(\mathcal{M})$  such that $\mathcal{M} \subset \mathcal{N}$ and that it is small enough so:

\begin{equation}
\label{eq:lowbound}
    E_{\mathcal{A}}(b) \leq E_{\mathcal{A}}(b^*) + \frac{\Delta_{\mathcal{N}}}{4} \ \forall b \in \mathcal{M}
\end{equation}

\noindent the existence of which comes from the continuity of $E_{\mathcal{A}}(b)$ in a  neighborhood of $b^*$ (the ESDF is always continuous everywhere), that $\Delta_{\mathcal{N}} > 0$ and that $b^*$ minimizes $E_{\mathcal{A}}$ in $\partial \mathcal{B}$. Thus,  due to \eqref{eq:DeltaN}:

\begin{equation}
\label{eq:uppbound}
    E_{\mathcal{A}}(b) \geq E_{\mathcal{A}}(b^*) + \Delta_{\mathcal{N}} \ \forall b \in \partial \mathcal{B}-\mathcal{N}
\end{equation}

Now, choose $h$ sufficiently small so that

\begin{equation}
\label{eq:maxbound}
    \max_{b \in \partial \mathcal{B}} |E_{\mathcal{A}}(b)-E^h_{\mathcal{A}}(b)| \leq \frac{\Delta_{\mathcal{N}}}{8}
\end{equation}

\noindent which is possible because $\Delta_{\mathcal{N}}>0$ and the third property in Definition \ref{def:hesdf}.  Now, note that :
\begin{eqnarray}
     &&I_1 = \int_{\partial\mathcal{B}-\mathcal{N}} e^{-E_{\mathcal{A}}^h(b)/h^2}dS = \nonumber \\
     && \int_{\partial\mathcal{B}-\mathcal{N}} e^{-E_{\mathcal{A}}(b)/h^2}e^{-(E_{\mathcal{A}}^h(b)-E_{\mathcal{A}}(b))/h^2}dS \underbrace{\leq}_{\mbox{eq. \eqref{eq:uppbound}}}  \nonumber \\
  && e^{-(E_{\mathcal{A}}(b^*)+\Delta_{\mathcal{N}})/h^2}\int_{\partial\mathcal{B}-\mathcal{N}} e^{-(E_{\mathcal{A}}^h(b)-E_{\mathcal{A}}(b))/h^2}dS \underbrace{\leq}_{\mbox{eq. \eqref{eq:maxbound}}} \nonumber \\
  && e^{-(E_{\mathcal{A}}(b^*)+\Delta_{\mathcal{N}})/h^2}e^{\Delta_{\mathcal{N}}/(8h^2)} \int_{\partial\mathcal{B}-\mathcal{M}}dS \underbrace{\leq}_{\partial \mathcal{B} - \mathcal{N} \subseteq \partial \mathcal{B}} \nonumber \\
  && e^{-(E_{\mathcal{A}}(b^*)+(7/8)\Delta_{\mathcal{N}})/h^2}\sigma(\partial \mathcal{B}).
\end{eqnarray}

And note that:

\begin{eqnarray}
   && I_2 = \int_{\mathcal{M}} e^{-E_{\mathcal{A}}^h(b)/h^2}dS = \nonumber \\
    && \int_{\mathcal{M}} e^{-E_{\mathcal{A}}(b)/h^2}e^{-(E_{\mathcal{A}}^h(b)-E_{\mathcal{A}}(b))/h^2}dS  \underbrace{\geq}_{\mbox{eq. \eqref{eq:lowbound}}} \nonumber \\
   && e^{-(E_{\mathcal{A}}(b^*)+(1/4)\Delta_{\mathcal{N}})/h^2} \int_{\mathcal{M}} e^{-(E_{\mathcal{A}}^h(b)-E_{\mathcal{A}}(b))/h^2}dS \underbrace{\geq}_{\mbox{eq. \eqref{eq:maxbound}}}  \nonumber \\
   && e^{-(E_{\mathcal{A}}(b^*)+(1/4)\Delta_{\mathcal{N}})/h^2}e^{-\Delta_{\mathcal{N}}/(8h^2)}\int_{\mathcal{M}}dS = \nonumber \\
   && e^{-(E_{\mathcal{A}}(b^*)+(3/8)\Delta_{\mathcal{N}})/h^2}\sigma(\mathcal{M}).
\end{eqnarray}

Let $I_3 = \int_{\partial \mathcal{B}} e^{-E_{\mathcal{A}}^h(b)/h^2}dS$ . Note that $I_3 \geq I_1+I_2$ since $\mathcal{M} \subset \mathcal{N}$. Thus, note that , $\mu_h(\partial \mathcal{B}-\mathcal{N}) = I_1/I_3  \leq I_1/(I_1+I_2) \leq I_1/I_2$ since $I_2 \geq 0$. Using the two inequalities for $I_1$ and $I_2$ we obtain the desired result with $C_{\mathcal{N}} \triangleq \sigma(\partial \mathcal{B})/\sigma(\mathcal{M})$.

\end{proof}

\textbf{Proof of Proposition 3}: 

Let $g_h(\beta) = n_{\mathcal{B}}(b)^{\top}\nabla E_{\mathcal{A}}^h(b)$ and $g^* = n_{\mathcal{B}}(b^*)^{\top}\nabla E_{\mathcal{A}}(b^*)$. Let $\mathcal{N} \subset \partial \mathcal{B}$ be an open subset containing $b^*$ in which $n_{\mathcal{B}}(b)$ is continuous (guaranteed to exist by the non-degeneracy conditions).  Set $\Delta g_h(b) = g_h(b)-g^*$.

Note that, using the triangle inequality and the fact that $\rho_{\mathcal{A},h}$ is a density and thus integrates to one:

\begin{eqnarray}
&& |N_{\mathcal{A},\mathcal{B}}^h{-}N_{\mathcal{A},\mathcal{B}}| \leq  \left| \int_{\partial \mathcal{B}} \rho_{\mathcal{A},h}(\beta)g_h(\beta)dS{-}g^*\right|{+}|d_{\mathcal{A}}^h{-}d_{\mathcal{A}}| \nonumber \\
    && =  \left| \int_{\partial \mathcal{B}} \rho_{\mathcal{A},h}(\beta)\Delta g_h(\beta)dS\right|+|d_{\mathcal{A}}^h-d_{\mathcal{A}}|. 
\end{eqnarray}

We will analyze the integral first. Let $M_h = \max_{b \in \partial \mathcal{B}} \|\nabla E_{\mathcal{A}}^h(b)\|$. In $\partial \mathcal{B}-\mathcal{N}$, we can provide the bound $\|\Delta g_h(b)\| \leq M_h + |g^*|$. 

In $\mathcal{N}$, we can use the fourth property in Definition \ref{def:hesdf}  (uniform convergence of the gradient of $E^h_{\mathcal{A}}(b)$) and  assure that for every $\eta > 0$ that we choose, there exists a $h$ small enough such that:

\begin{equation}
\label{eq:Deltagh_bound}
    \sup_{b \in \mathcal{N}} |\Delta g_h(b)| \leq \eta.
\end{equation}

Pick $h$ small enough so Lemma \ref{lemma:lemma3} and \eqref{eq:Deltagh_bound} holds for the chosen set $\mathcal{N}$. Thus, using the triangle inequality and  the definition of $\mu_h$ in \eqref{eq:muh}:

\begin{eqnarray}
 && \left| \int_{\partial \mathcal{B}} \rho_{\mathcal{A},h}(\beta)g_h(\beta)dS-g^*\right| \leq \nonumber \\ 
 &&  \leq \left| \int_{\partial \mathcal{B}-\mathcal{N}} \rho_{\mathcal{A},h}(\beta)\Delta g_h(\beta)dS\right| + \left| \int_{\mathcal{N}} \rho_{\mathcal{A},h}(\beta)\Delta g_h(\beta)dS\right|  \nonumber \\
 && \leq \eta \mu_h(\mathcal{N}) + \big(M_h+|g^*|\big)\mu_h(\partial \mathcal{B}-\mathcal{N}) \leq \nonumber \\
&& \eta \mu_h(\mathcal{N}) + \big(M_h+|g^*|\big)C_{\mathcal{N}}e^{-\Delta_{\mathcal{N}}/(2h^2)}.
\end{eqnarray}

The upper bound of the error can go to $0$ when $h \rightarrow +0$. This is because $\eta$ can be made arbirtrarily small, $\mu_h(\mathcal{N}) \leq \mu_h(\partial \mathcal{B}) = 1$, and the fifth property in Definition \ref{def:hesdf} guarantes that $M_h e^{-\Delta_{\mathcal{N}}/(2h^2)} \rightarrow 0$. Finally, the second propery in Definition \ref{def:hesdf} guarantees that  $|d_{\mathcal{A}}^h-d_{\mathcal{A}}|$ as $h \rightarrow +0$. And the proof is complete.

$\blacksquare$

\textbf{Proof of Proposition 4}: Note the $f(t)$ can be written as $d^h_{\mathscr{a}} - f_{num}(t)/f_{den}(t)$, in which $f_{den}(t) > 0$ for all $t$. Thus, differentiability follows if the differentiability of $f_{num}(t)$ and $f_{den}(t)$ are established.

We will prove for $f_{num}(t)$. The proof for the other follows similarly. The major step is to rewrite the integral using the formulas for the rigid motion.

Now, define  $T_{\mathcal{A}}^{-1}(y,t) = Q_{\mathcal{A}}(t)^{\top}(y-s_{\mathcal{A}}(t))$, $T_{\mathcal{B}}(y,t) = Q_{\mathcal{B}}(t)y+s_{\mathcal{B}}(t)$ and $T_{\mathcal{A}}^{\mathcal{B}}(\beta,t) = T_{\mathcal{A}}^{-1}(T_{\mathcal{B}}(\beta',t),t)$.  Let $F(\beta',t)$ be defined as
\begin{equation}
     e^{-\frac{E^h_{\mathscr{a}}\big(T_{\mathcal{A}}^{\mathcal{B}}(\beta',t)\big)}{h^2}} n_{\mathscr{b}}(\beta')^{\top}Q_{\mathcal{B}}(t)^{\top}Q_{\mathcal{A}}(t)\nabla E^h_{\mathscr{a}}\big(T_{\mathcal{A}}^{\mathcal{B}}(\beta',t)\big).
\end{equation}
Using \eqref{eq:Dh_transf}  and the change of variable $\beta = T_{\mathcal{B}}(\beta',t)$, we can write
\begin{equation}
    f_{num}(t) = \int_{\partial \mathscr{b}} F(\beta',t)dS.
\end{equation}

This is $k$ times differentiable as long as $F(\beta,t)$ is $k$ times differentiable in $t$. Since $E^h_{\mathscr{a}}$ is infinitely differentiable, the limit of differentiability is on $Q_{\mathcal{A}}(t)$, $Q_{\mathcal{B}}(t)$, $s_{\mathcal{A}}(t)$ and $s_{\mathcal{B}}(t)$.
$\blacksquare$
\bibliographystyle{ieeetr}  
\bibliography{bibl}
\end{document}